\documentclass[10pt,twocolumn,letterpaper]{article}

\usepackage[pagenumbers]{cvpr} 

\newcommand{\fref}[1]{Figure~\ref{#1}}

\newcommand{\tref}[1]{Table~\ref{#1}}
\newcommand{\sref}[1]{Section~\ref{#1}}
\newcommand{\aref}[1]{Appendix~\ref{#1}}

\newcommand{\mingi}[1]{\textcolor{black}{#1}}

\newcommand{\js}[1]{\textcolor{black}{#1}}

\newcommand{\vv}[1]{\mathbf{v}}
\newcommand{\vvt}[1]{\mathbf{v}_{t}}
\newcommand{\vx}[1]{\mathbf{x}}
\newcommand{\vz}[1]{\mathbf{z}}
\newcommand{\vzt}[1]{\mathbf{z}_{t}}

\newcommand{\ours}[1]{FlowBlending}

\def\maketitlesupplementary{
    \clearpage
    \onecolumn
    \begin{center}
        \Large
        \textbf{FlowBlending: Stage-Aware Multi-Model Sampling for Fast and High-Fidelity Video Generation}\\[0.5em]
        \normalsize Supplementary Material
    \end{center}
    \vspace{1.0em}
}

\definecolor{cvprblue}{rgb}{0.21,0.49,0.74}
\usepackage[pagebackref,breaklinks,colorlinks,allcolors=cvprblue]{hyperref}
\usepackage{kotex}
\usepackage{cuted}
\usepackage{caption}
\usepackage{multirow}
\usepackage{graphicx}
\usepackage{subcaption}
\usepackage{algorithm}
\usepackage{algorithmic}

\def\paperID{} 
\def\confName{CVPR}
\def\confYear{2026}

\title{FlowBlending: Stage-Aware Multi-Model Sampling \\ for Fast and High-Fidelity Video Generation}

\author{
Jibin Song, Mingi Kwon, Jaeseok Jeong, Youngjung Uh\thanks{Corresponding author.} \\
Yonsei University \\
{\tt\small
\{sjbpsh1, kwonmingi, jete\_jeong, yj.uh\}@yonsei.ac.kr}
}

\begin{document}

\maketitle

\begin{strip}
  \centering
  \vspace{-5em}
  \includegraphics[width=\textwidth]{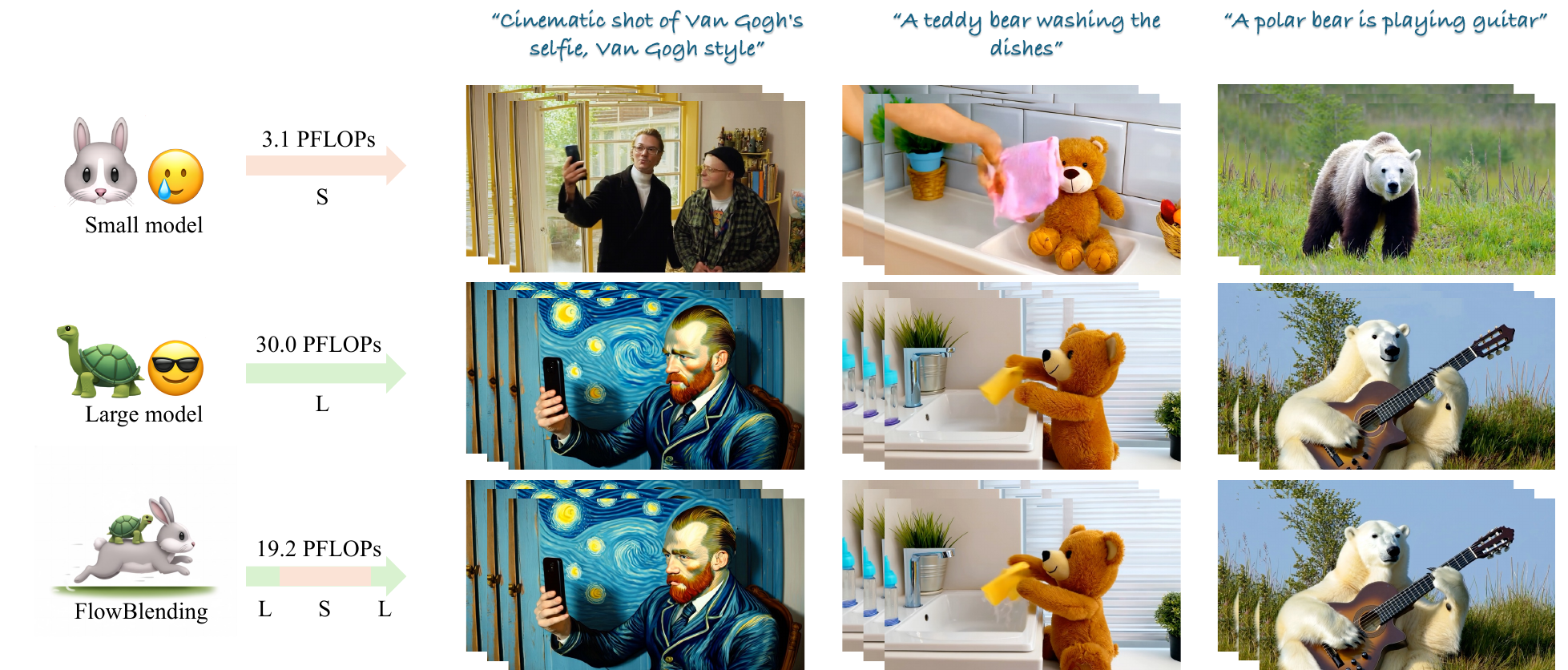}
  \vspace{-2em}
  \captionof{figure}{\textbf{Overview of \ours{}.}
  The videos in each column are generated from the same initial noise and text prompt, but with different model allocation strategies. \ours{} assigns a large model to process early and late denoising stages---establishing global structure and refining details, respectively--- and assigns a small model to process intermediate denoising stages, where velocity divergence between the two models is minimal. 
  This approach \textit{preserves visual fidelity} of the large model while \textit{reducing computation.}}
  \label{fig:concept}
\end{strip}

\begin{abstract}

In this work, we show that the impact of model capacity varies across timesteps: it is crucial for the early and late stages but largely negligible during the intermediate stage.
Accordingly, we propose \ours{}, a stage-aware multi-model sampling strategy that employs a large model and a small model at capacity-sensitive stages and intermediate stages, respectively. We further introduce simple criteria to choose stage boundaries and provide a velocity-divergence analysis as an effective proxy for identifying capacity-sensitive regions.
Across LTX-Video (2B/13B) and WAN 2.1 (1.3B/14B), \ours{} achieves up to 1.65× faster inference with 57.35\% fewer FLOPs, while maintaining the visual fidelity, temporal coherence, and semantic alignment of the large models. \ours{} is also compatible with existing sampling-acceleration techniques, enabling up to 2× additional speedup.
Project page is available at: \href{https://jibin86.github.io/flowblending_project_page/}{https://jibin86.github.io/flowblending\_project\_page}

\end{abstract}    
\section{Introduction}
\label{sec:intro}
Recent advances in diffusion-based video generation have significantly improved visual fidelity and temporal coherence \cite{ho2022video, hong2022cogvideo, blattmann2023stable,  chen2023videocrafter1, blattmann2023videoldm, wang2023modelscope, brooks2024video, hacohen2024ltx, wan2025wan, jin2024pyramidal, kwon2024harivo}.
However, these gains come with substantial computational cost due to the iterative denoising process and the increasing size of modern video diffusion models.
While sampling acceleration has been extensively explored in text-to-image diffusion, progress on accelerating video diffusion has been limited.

Existing methods for accelerating diffusion sampling generally fall into two categories.
One line of work accelerates sampling by sampling-step reduction algorithm, for example through improved numerical solvers or trajectory approximations \cite{nichol2021improved, karras2022elucidating, song2022ddim, lu2022dpm, Lu_2025}.
Another direction distills the diffusion process into a smaller number of forward passes, aiming to match full-sampling performance with significantly fewer steps \cite{frans2024one, yin2024one, kim2024simple, mei2024codi, dao2024swiftbrush, salimans2022progressive, li2023snapfusion, xu2024ufogen, sauer2024adversarial, sauer2024fast, hsiao2024plug}.
However, both strategies predominantly assume that \emph{all timesteps require the same model capacity}, either by applying a single model uniformly across the denoising schedule or by compressing the entire model into a single distilled variant.

This leads to a fundamental question:
\emph{Do we truly need a large model for every diffusion step?}
Notably, many recent video diffusion models are released in multiple capacity variants, e.g., LTX-Video\cite{hacohen2024ltx} (2B / 13B) and WAN~2.1\cite{wan2025wan} (1.3B / 14B).
Small models are significantly faster, but their visual quality is typically inferior, and they often fail to preserve semantic details.
As illustrated in \fref{fig:concept}, the WAN~2.1 small model (3.1 PFLOPs) struggles to accurately follow text prompts and often produces distorted or collapsed objects, despite its substantial efficiency advantage.
In contrast, the large model, though computationally expensive, generates temporally coherent and semantically faithful videos.
This disparity between large and small models highlights an opportunity: instead of uniformly applying high capacity across all timesteps, one may allocate the large model only where it provides the greatest benefit.

In this work, we make a key empirical observation inspired by the characteristics of the sampling process in video diffusion:
\textbf{model capacity is not uniformly important across timesteps.}
By evaluating the large and small models across multiple schedules, we find that:
The early denoising stage influences global structure and motion, where the large model capacity yields noticeably better structure and semantic alignment than the small model.
In addition, the late denoising stage refines high-frequency details and remove artifacts, again benefiting from the expressiveness of the larger model.
In contrast, we demonstrate that the intermediate denoising stage admits substantial capacity reduction, as the small model yields outputs nearly identical to those of the large model.

These findings indicate that certain stages of the sampling process are substantially more capacity-sensitive than the others.
Building on this insight, we propose \ours{}, where the large model is used only in the early and late denoising stages, while the small model handles the majority of intermediate denoising stage.
As shown in \fref{fig:concept}, this approach preserves the visual quality of the large model while reducing computational cost a lot.
Notably, our method requires no additional training, distillation, or architectural modification, and is complementary to existing acceleration techniques.

Furthermore, to identify when each model should be used, we introduce simple and practical strategies: semantic similarity between latents and a quantitative indicator of fine-detail quality. These strategies enable efficient capacity allocation while maintaining generation quality that is nearly indistinguishable from the large model.

In addition, we provide extensive experiments together with a velocity divergence analysis, which offers further insight into the sampling process and suggests a principled way to identify the early and late stage boundaries.

Across two open-source video diffusion models, LTX-Video (2B / 13B) and WAN 2.1 (1.3B / 14B), our stage-aware multi-model sampling achieves up to \textbf{1.65× faster} inference with \textbf{57.35\% FLOPs}
, while preserving the performance of the large model in visual fidelity, temporal coherence, and semantic alignment. In addition, our approach is orthogonal to existing acceleration methods, allowing up to an additional 50\% FLOPs reduction when combined with complementary techniques.

\section{Related work}
\label{sec:related}

\paragraph{Background: flow matching}
Within the great success of diffusion models \citep{ho2020ddpm,song2021scorebasedgenerativemodelingstochastic}, flow matching \citep{lipman2022flow,lipman2024flowmatchingguidecode,tong2024improvinggeneralizingflowbasedgenerative} has emerged as a widely adopted framework for modern generative modeling \citep{esser2024scaling, labs2025flux, hacohen2024ltx, wan2025wan}. Flow matching transfers the source distribution $p_{0}$ (e.g., Gaussian noise) to the target distribution $p_{1}$ (e.g., data distribution) by learning a velocity field $\vvt{}(\vx{};\theta)$ with a neural network $\theta$. According to conditional flow matching (CFM), an intermediate latent $\vz{}_{t}= (1-t)\vz{}_0 + t\vz{}_{1}$ is formed at each timestep $t$ and the network $\theta$ is trained using the optimal transport CFM (OT-CFM) loss:
\begin{equation}
\resizebox{\columnwidth}{!}{$
\mathcal{L}_{\text{CFM}}(\theta)
= \mathbb{E}_{t,\, q(\vz{}_1),\, p(\vz{}_0)}
\left\|
\vv{}_t\left((1 - t)\vz{}_0 + t\,\vz{}_1\right)
- (\vz{}_1 - \vz{}_0)
\right\|^2.
$}
\end{equation}
The learned $\vvt{}(\cdot;\theta)$ progressively transfers the randomly sampled $z_{0} \sim N(0,\text{I})$ into $z_1$ by solving ordinary differential equation (ODE) with the number of function evaluations (NFE) where flow step $t \in [0,1]$.

\paragraph{Acceleration of diffusion models}

Diffusion models inherently suffer from high computational cost due to the large number of denoising steps required during sampling process.
One line of work aims to mitigate this by employing more efficient numerical solvers, which substantially reduce the number of function evaluations (NFE) without additional training \cite{nichol2021improved, karras2022elucidating, song2022ddim, lu2022dpm, Lu_2025}.
Another major direction focuses on step distillation \cite{frans2024one, yin2024one, kim2024simple, mei2024codi, dao2024swiftbrush, salimans2022progressive, li2023snapfusion, xu2024ufogen, sauer2024adversarial, sauer2024fast, hsiao2024plug}, which compresses multi-step sampling into a smaller number of steps.
Recently, several extensions have been proposed for video diffusion models as well \cite{zhang2025accvideo, ding2025efficient, wu2025taming, zhang2024sf, wu2025snapgen}, though these approaches typically require costly retraining.
These two lines of research are largely orthogonal to our approach, as we focus on switching between models of different capacities during the sampling process rather than reducing the number of steps or retraining the model.

\paragraph{Multi-model sampling} \citet{yang2023denoising} showed that model capacity needs differ across timesteps in image diffusion, while \citet{liu2023oms} and \citet{pan2024t} proposed mixing or allocating image models of varying sizes to different steps. However, their image-based analysis does not extend to video diffusion, where we find the opposite trend: \textit{large models are crucial in the early stage to establish structure and coherent motion.}

\section{Stage-aware multi-model sampling}

A key challenge in accelerating video diffusion is to retain the generative performance of a high-capacity model while substantially reducing computational cost.
Instead of modifying the sampling algorithm or retraining via distillation, we aim to reuse the existing large and small models as-is, allocating them across the sampling process.
In other words, we frame this as a capacity allocation problem:

\textbf{How can we preserve the quality of the large model while using the small model wherever possible?}

Although the large model consistently yields higher fidelity, using it across all timesteps is prohibitively expensive.
If we can identify the denoising timesteps where the small model’s updates closely match those of the large model, we can safely replace the large model during those timesteps without degrading structure, motion dynamics, or identity consistency.

In the following sections, we analyze how model capacity contributes differently across the sampling process.
In \sref{sec:early}, we demonstrate that the early stage primarily governs global structure and coarse motion, where the expressive capacity of the model plays a critical role.
Then, in \sref{sec:late}, we show that the late stage is responsible for refining high-frequency details and resolving artifacts, again benefiting from the representational strength of the larger model. Finally, in \sref{sec:flowblending}, we introduce our method that leverages both characteristics.

\subsection{Early structure formation}
\label{sec:early}

Our goal is to replace portions of the sampling process with the small model while retaining the large model’s performance.
We posit that key quality factors, such as visual fidelity, temporal coherence, and semantic alignment, are largely shaped in the early stage of the denoising process.
We verify it in the video diffusion model via a simple but illustrative ablation.

\begin{figure}[t]
  \centering
  \includegraphics[width=\linewidth]{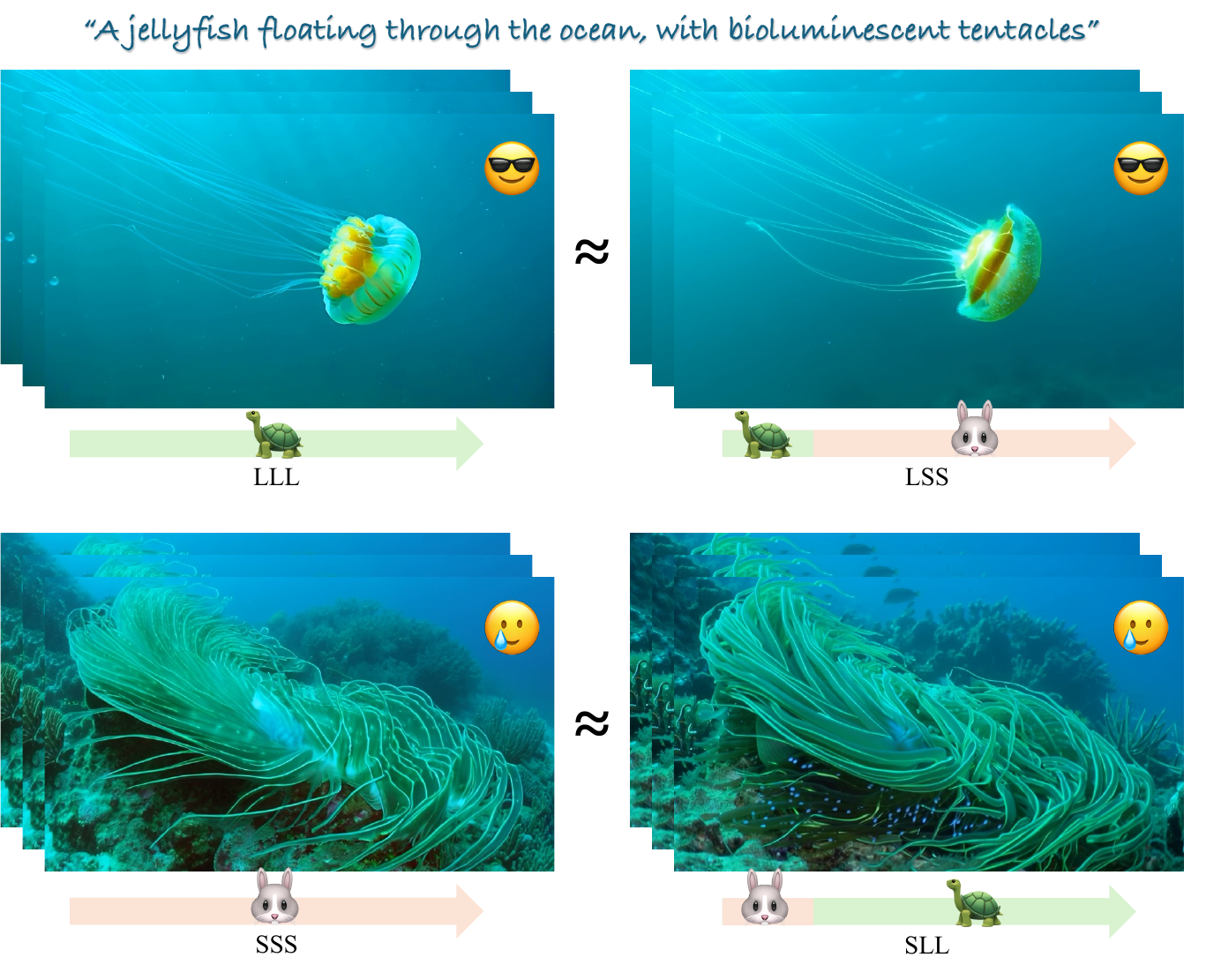}
  \vspace{-2em}
  \caption{
  \textbf{Effect of model capacity during early denoising stage.}
  Comparison of WAN-2.1 sampling schedules (L = 14B, S = 1.3B).  
  LSS (large only in early steps) closely matches LLL (large-only) in structure and motion,  
  while SSS (small-only) exhibits temporal inconsistency and semantic misalignment. SLL (small only in early steps) likewise produces structure and motion patterns highly similar to SSS.
  This shows that the early stages are crucial for establishing global semantic and structural attributes.}
  \vspace{-1.5em}
  \label{fig:early}
\end{figure}

\fref{fig:early} compares four sampling schedules for WAN-2.1 (14B model as L, 1.3B as S):  
LLL (large-only), LSS (large$\rightarrow$small), SLL (small$\rightarrow$large), and SSS (small-only),  
where each letter denotes which model is used for a predefined segment of the timestep schedule.

As expected, LLL yields coherent global structure and strong alignment with the prompt.
Remarkably, LSS, where the large model is used only during the earliest timesteps, produces nearly identical structure, motion coherence, and semantic consistency to LLL.
In contrast, SSS fails to generate videos consistent with the text prompt, often producing drifting motion or incorrect object identity.
Even SLL, which applies the large model after the initial steps, behaves similarly to SSS and struggles to recover correct semantics once the early structure formation has been misaligned.

These results indicate that \textbf{the early stage is capacity-sensitive}:
The large model is critical for establishing coarse structure-level and motion-level attributes.
Once the large model establishes the coarse structure, the small model can successfully take over the denoising process with minimal degradation in perceptual quality. We further provide the quantitative experiment in \sref{sec:early_stage_quantitative}.

\subsection{Late refinement}
\label{sec:late}

While LSS preserves global structure, motion, and semantics, we find that it may introduce subtle artifacts during the later denoising stage.
These artifacts typically appear as spatial distortions or temporal flicker, which do not affect the overall structure but noticeably degrade perceptual quality.

\begin{figure}[t]
  \centering
  \includegraphics[width=\linewidth]{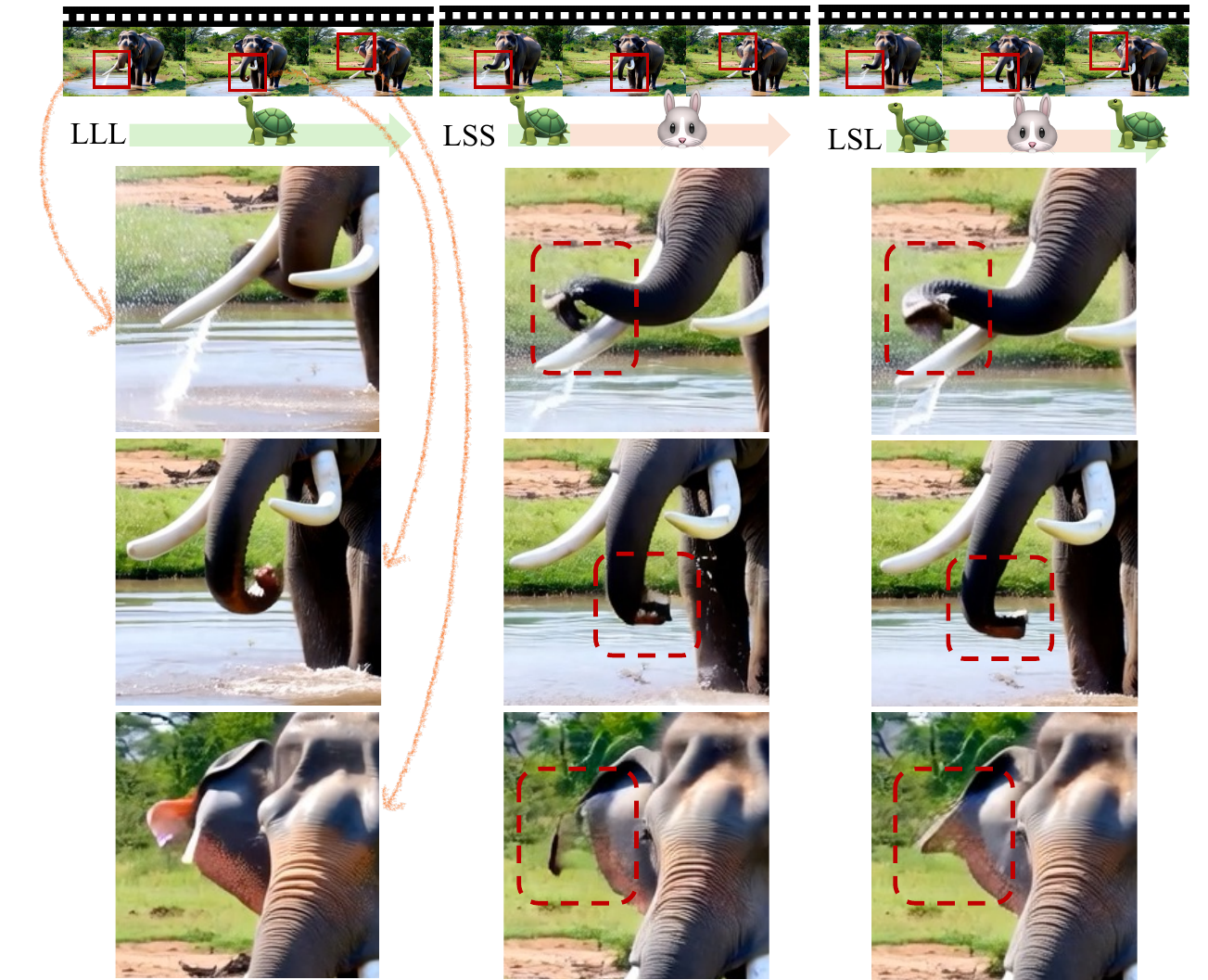}
  \vspace{-1em}
  \caption{
  \textbf{Late stage ablation: LLL vs.\ LSS vs.\ LSL.}
  LSS preserves global structure similar to LLL but exhibits some artifacts.  
  Reintroducing the large model only during the late stage (LSL) restores detail and reduces flicker,  
  demonstrating that the late denoising stage is capacity-sensitive.
  Notably, the LSL schedule attains quality nearly indistinguishable from LLL while retaining the efficiency benefits of using the small model for most of the trajectory. Please zoom in to view the figures in detail.
  }
  \vspace{-1.5em}
  \label{fig:late}
\end{figure}

\fref{fig:late} illustrates this effect by comparing LLL (large-only), LSS (large$\rightarrow$small), and LSL (large$\rightarrow$small$\rightarrow$large).
Here, LSS uses the large model only during the early stage, while LSL reintroduces the large model exclusively in the final few timesteps.

As discussed in the previous subsection, LSS remains visually close to LLL in terms of coarse structure and motion.
However, it fails to resolve high-frequency artifacts.
In contrast, LSL successfully suppresses late stage distortions and restores fine detail.
This indicates that the role of the large model in the late stage is not structure formation (which is already established early), but \emph{artifact correction and detail refinement}.
We therefore argue that \textbf{the late stage is capacity-sensitive once again}. We further provide the quantitative experiment in \sref{sec:late_stage_quantitative}.

\subsection{Stage-aware multi-model sampling}
\label{sec:flowblending}
Based on these observations, we propose \textbf{FlowBlending (LSL)}, which allocates model capacity according to the sensitivity of each stage in the denoising process.
The large model is used only during the early and late stages, where global structure formation and detail refinement are critical, while the small model handles the capacity-tolerant intermediate stage.
This simple yet effective scheduling achieves near-large-model quality with reduced computation.

In the next section \sref{sec:quantitative}, we provide a quantitative evaluation of this strategy, analyzing its trade-offs across model blendings.
We further describe how we determine the optimal boundaries of the early and late stages, inspired by the analysis between large and small models.

\section{Experiment}
\label{sec:quantitative}
\subsection{Experimental setup} 
We evaluate the proposed sampling schedule on two representative open-source video diffusion models: LTX-Video \cite{hacohen2024ltx} (2B / 13B) and WAN 2.1 \cite{wan2025wan} (1.3B / 14B).
Evaluations are conducted on the PVD \cite{bolya2025perception} and VBench \cite{huang2024vbench}.
We report FID \cite{fid} and FVD \cite{fvd} using 284 generated samples, and four VBench metrics, Aesthetic Quality, Background Consistency, Subject Consistency, and Motion Smoothness, using 355 generated samples.
We follow the default settings from each official repository.
To quantify computational efficiency, we report the runtime and FLOPs of DiT blocks per generated video. Runtime for LTX-Video is measured on a NVIDIA A6000 GPU, and runtime for WAN 2.1 is measured on a NVIDIA A100 GPU. Please refer to \aref{appdixsec:exp_setting} for more details.

\subsection{Analysis of the early stage}
\label{sec:early_stage_quantitative}
In \sref{sec:early}, we show that applying the large model only in the early stage is sufficient to establish the global structure and motion, after which the small model can take over the remaining denoising steps.

To quantitatively support the observation, we measure the similarity between each sampling schedule and the large-only baseline (LLL) using four metrics: (i) DINO \cite{caron2021emerging} and CLIP \cite{radford2021learning} image-embedding similarity for semantic consistency, and (ii) LPIPS \cite{zhang2018unreasonable} and PSNR for low-level similarity, averaged across all frames and 355 generated videos.
As shown in \tref{tab:similarity}, LSS remains substantially closer to LLL across both semantic and low-level similarity metrics, confirming that invoking the large model only during the early stage is sufficient to preserve global structure and text-aligned semantics.
In contrast, both SLL and SSS exhibit significantly lower similarity.
Notably, SLL performs comparably to SSS despite using the large model in the latter part of sampling. This result indicates that if the early structure formation is misaligned, the subsequent steps cannot recover the global structure or the semantic alignment to the prompt, even when a high-capacity model is used later in the denoising process. 

These findings suggest that \textbf{the early stage is capacity-sensitive}, where employing the large model is crucial for establishing stable global structure and semantic alignment.
For full experimental details, please refer to \aref{appdixsec:more_abl_early}.

\begin{table}[]
\resizebox{\columnwidth}{!}{%
\begin{tabular}{lcccc}
\toprule
\textbf{Schedule}             & \textbf{DINO Sim}$\uparrow$ & \textbf{CLIP Sim}$\uparrow$ & \textbf{LPIPS}$\downarrow$ & \textbf{PSNR}$\uparrow$ \\ \midrule
LLL (Large-only)              & 100.00              & 100.00            & -              & -             \\ \midrule
LSS (Early-large, then small) & \textbf{95.74}               & \textbf{96.97}             & \textbf{2.76}           & \textbf{24.30}         \\
SLL (Early-small, then large) & 65.58               & 81.10             & 30.49          & 12.62         \\
SSS (Small-only)              & 65.01               & 80.87             & 30.54          & 12.59         \\ \bottomrule
\end{tabular}%
}
\caption{
\textbf{Similarity to large-model baseline.}
We report cosine similarity of DINO embeddings and CLIP embeddings, LPIPS, and PSNR averaged per-frame across the evaluation set.
LSS remains close to LLL, indicating that early use of the large model preserves global semantics, 
while SSS diverges significantly.
}
\vspace{-1.5em}
\label{tab:similarity}
\end{table}

\begin{figure}[t]
  \centering
  \includegraphics[width=\linewidth]{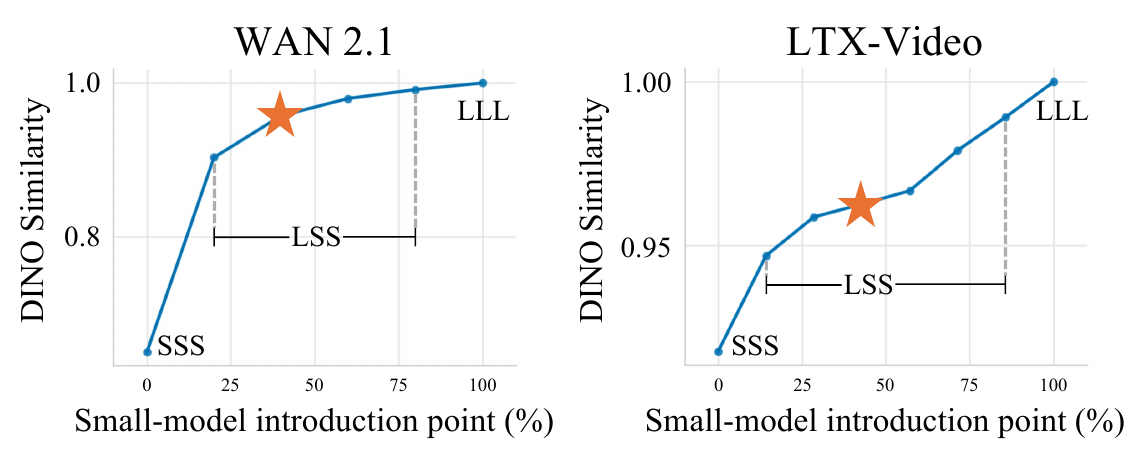}
  \vspace{-2em}
  \caption{
  \textbf{DINO-based identification of the early stage boundary.}
We measure the frame-wise DINO similarity between schedules that switch from the large to the small model ($L \rightarrow S$) at different early stage boundaries and the large-only baseline (LLL).
A sharp decline in similarity emerges beyond a specific point in the curve.
Boundaries chosen just before this drop (typically above $\sim$96\% similarity) preserve global structure and motion.
  }
  \vspace{-1.5em}
  \label{fig:dino_graph}
\end{figure}

\paragraph{Boundary of the early stage}
\fref{fig:dino_graph} extends the DINO-similarity experiment by varying the early stage boundary at which the schedule transitions from the large model to the small model ($L\rightarrow{}S$ on the LSS schedule).
We first observe that the DINO similarity curve exhibits a sharp drop after a certain point; before this drop, the similarity to the large-only baseline remains consistently high.
Motivated by this structure in the graph, we select the early-stage boundary just before the curve begins to decline.
Empirically, all such candidates fall above roughly $96\%$ similarity, and schedules using these boundaries generate videos whose global structure and motion are nearly indistinguishable from the large-only baseline.
This indicates that maintaining similarity above this threshold is sufficient to preserve high-level fidelity while enabling acceleration.

\subsection{Analysis of the late stage}
\label{sec:late_stage_quantitative}
In \sref{sec:late}, we show that applying the large model at the late stage refines details and reduces artifacts.
\tref{tab:overall_quan} quantitatively supports it through FID results:
The LSL schedule consistently achieves lower FID than LSS, indicating that late stage large-model updates improve visual fidelity.

Taken together, these results lead to our core principle:
\emph{Use the large model early to establish global structure, and use it again at the end to refine details.}

\begin{figure}[t]
  \centering
  \includegraphics[width=\linewidth]{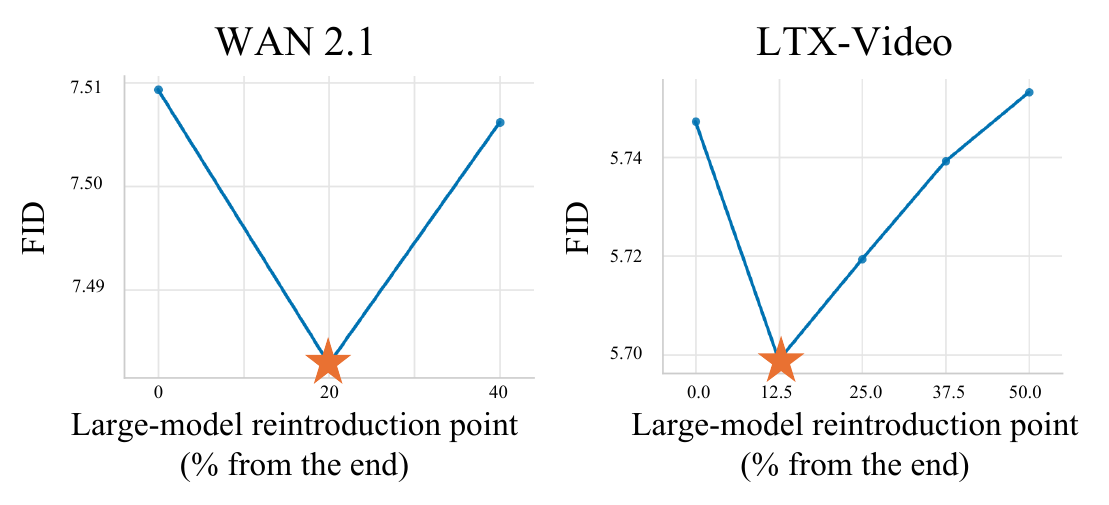}
  \vspace{-2em}
  \caption{
  \textbf{FID-based identification of the late stage boundary.}
  We fix the early stage boundary identified in \sref{sec:early_stage_quantitative} and vary only the late stage boundary. 
The resulting FID curve exhibits a V-shape, where the minimum corresponds to the optimal late stage boundary used in our LSL schedule. 
This trend consistently appears in both WAN and LTX-Video, demonstrating that a properly chosen late boundary yields the best sweet spot with detail refinement and artifact suppression.
  }
  \vspace{-1em}
  \label{fig:fid_curve}
\end{figure}

\paragraph{Boundary of the late stage.}
We observe that the FID curve exhibits a V-shape in \fref{fig:fid_curve}.
Interestingly, for both WAN and LTX-Video, choosing an appropriate late stage boundary consistently lowers the FID.
Moreover, when the early stage boundary is properly selected, all LSL schedules produce outcomes that closely match those of the large-only baseline (LLL).

A notable finding is that incorporating the small model during the intermediate stage can make the generated video appear more similar to real footage. While LLL produces high-quality results overall, it occasionally yields overly smooth surfaces, whereas well-configured LSL schedules generate objects with more natural and realistic textures. We speculate that a moderate degree of “noisiness” introduced by the small model may contribute to this increased realism.

However, excessive reliance on the small model leads to artifacts or temporal flicker, and the FID curve increases once the proportion of small model usage surpasses a certain threshold. The observed V-shape FID plot in \fref{fig:fid_curve} reflects this behavior. Our experiments reveal a clear sweet spot: moderate use of the small model during the intermediate stage, followed by large model refinement at the end, enhances fine-grained texture without introducing artifacts. We further argue that this sweet spot can be identified through artifact-sensitive and pixel-level detail metrics such as FID. Additional qualitative results are provided in \aref{appendix:late_stage_ablation_qual}.

\begin{table*}[t]
    \centering
    \small
    \caption{
    \textbf{Comparison of video quality and efficiency across sampling schedules.}
    LSL achieves quality comparable to the large-only baseline (LLL), while reducing runtime and FLOPs.
    LSS preserves global structure but leaves late stage artifacts, whereas SSS degrades across most metrics.
    }
    \label{tab:overall_quan}

\begin{subtable}{\textwidth}
\begin{tabular*}{\textwidth}{@{\extracolsep{\fill}}lcccccccc}
\toprule
\textbf{Schedule} & \textbf{FID} $\downarrow$ & \textbf{FVD} $\downarrow$ &
\textbf{Aesthetic} $\uparrow$ & \textbf{Background} $\uparrow$ &
\textbf{Subject} $\uparrow$ & \textbf{Motion} $\uparrow$
& \textbf{Runtime} $\downarrow$ & \textbf{TFLOPs} $\downarrow$ \\
\midrule
LLL (Large-only) & \underline{5.73} & 834.26 & \textbf{45.43} & 93.79 & \textbf{90.96} & 97.56
& 49.73{\tiny$\pm$0.31} & 3496 \\
LSL (Stage-aware, Ours) & \textbf{5.70} & \textbf{752.07} & \underline{44.52} & \textbf{93.96} & \underline{90.79} & \textbf{97.74}
& 30.18{\tiny$\pm$0.04} & 2005 \\
LSS (Early-large) & 5.75 & \underline{759.39} & 44.25 & \underline{93.89} & 90.66 & \underline{97.63}
& 25.44{\tiny$\pm$0.07} & 1632 \\
SSS (Small-only) & 6.28 & 951.79 & 43.14 & 91.70 & 86.23 & 96.02
& 10.69{\tiny$\pm$0.01} & 514 \\
\bottomrule
\end{tabular*}
\caption{
\textbf{LTX-Video}. Runtime is measured on a NVIDIA A6000 GPU.
}
\label{tab:quality_ltx}
\end{subtable}

\begin{subtable}{\textwidth}
\centering
\small
\begin{tabular*}{\textwidth}{@{\extracolsep{\fill}}lcccccccc}
\toprule
\textbf{Schedule} & \textbf{FID} $\downarrow$ & \textbf{FVD} $\downarrow$ &
\textbf{Aesthetic} $\uparrow$ & \textbf{Background} $\uparrow$ &
\textbf{Subject} $\uparrow$ & \textbf{Motion} $\uparrow$
& \textbf{Runtime} $\downarrow$ & \textbf{TFLOPs} $\downarrow$ \\
\midrule
LLL (Large-only)        & 7.55 & 2,831.94 & \textbf{57.56} & \underline{96.21} & \underline{94.61} & \underline{98.10}
& 653.12{\tiny$\pm$0.80} & 29950 \\
LSL (Stage-aware, Ours) & \textbf{7.48} & \underline{2,752.17} & \underline{57.42} & \textbf{96.29} & \textbf{94.61} & \textbf{98.11}
& 439.36 {\tiny$\pm$0.56} & 19222 \\
LSS (Early-large)       & \underline{7.51} & 2,803.48 & 57.27 & 96.11 & 94.47 & 98.10
& 335.20 {\tiny$\pm$0.44} & 13858 \\
SSS (Small-only)        & 7.55 & \textbf{2,556.55} & 56.21 & 95.71 & 94.15 & 97.93
& 124.05 {\tiny$\pm$0.47} & 3129 \\
\bottomrule
\end{tabular*}
\caption{
\textbf{WAN 2.1}. Runtime is measured on a NVIDIA A100 GPU.
}
\vspace{-1em}
\label{tab:quality_wan}
\end{subtable}
\end{table*}

\subsection{Quantitative results}

\tref{tab:overall_quan} summarizes the quantitative results in schedules across two models.
Our proposed \textbf{LSL} schedule achieves FID, FVD, Aesthetic, Background, Subject, and Motion scores that are nearly indistinguishable from the large-only baseline (LLL), while accelerating inference by up to $1.65\times$ with 57.35\% FLOPs.
Compared to the small-only schedule (SSS), LSL clearly preserves the performance characteristics of LLL, demonstrating that selectively invoking the large model at key stages is sufficient to maintain overall quality.

In contrast, the early-large schedule (LSS) successfully preserves global structure but fails to resolve high-frequency artifacts, resulting in consistently lower scores across several metrics. This behavior aligns with our analysis in \fref{fig:late}, where late stage model capacity is shown to be critical for detail refinement.

These results reinforce the central claim of this work:
\textbf{Video quality does not require high model capacity at every step; invoking the large model only at structurally and detail-critical stages preserves fidelity while reducing computation.}

\section{\mingi{Extended analyses}}
\label{sec:ablation}

In \sref{sec:divergence_analysis}, we examine the distribution of velocity divergence between the large and small model velocities and analyze how its behavior aligns with the identified boundary regions.
In \sref{sec:diverse_configurations}, we validate the proposed stage-aware schedule across a wide range of configurations to demonstrate that our boundary choice is appropriate.

\subsection{Velocity divergence across sampling steps}
\label{sec:divergence_analysis}

A natural way to determine when to use the large or small model is to measure how differently they update the latent state during denoising.
To this end, we compare the velocity fields predicted by the large and small models at each timestep $t$.
Given a latent $\vz{}_t$, let $\vv{}^{(L)}_t$ and $\vv{}^{(S)}_t$ denote the velocities predicted by the large (L) and small (S) models, respectively.
We compute two metrics:
$\text{cosine\_dist}(t) = 1 - \cos\big(\vv{}^{(L)}_t, \vv{}^{(S)}_t\big)$, 
$\ell_2\text{\_dist}(t)$ $ = \|\vv{}^{(L)}_t - \vv{}^{(S)}_t\|_2$.

\begin{figure}[t]
  \centering
  \includegraphics[width=\linewidth]{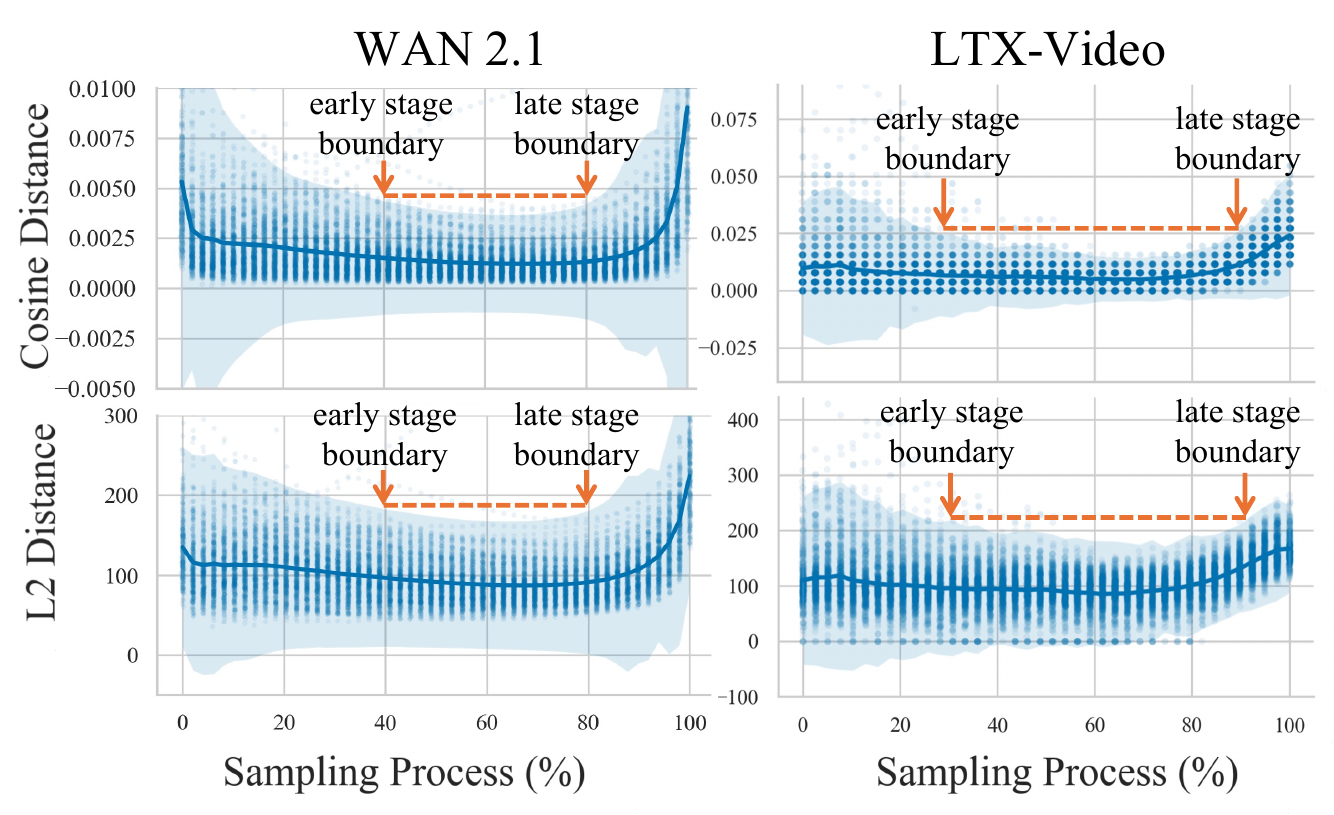}
  \vspace{-2em}
  \caption{
  \textbf{Velocity divergence across diffusion steps.}
  We compute cosine and $\ell_2$ distances between velocities $\vv{}^{(L)}_t$ and $\vv{}^{(S)}_t$ across timesteps.  
  Divergence is lowest in the intermediate stage, indicating that the small model can reliably denoise there.
  Early stage exhibits high variance (structure formation), while late stage shows high mean divergence (detail refinement).
  This supports allocating the large model to the early and late stages.
  }
  \vspace{-1em}
  \label{fig:velocity_divergence}
\end{figure}

We estimate these quantities by sampling from random noise using the large model, while feeding the same intermediate latents through the small model.
As shown in \fref{fig:velocity_divergence}, the divergence curve consistently follows a U-shaped pattern across the sampling process. The intermediate timesteps show low divergence, indicating that the small model and large model predict nearly identical update directions in this region.
In contrast:
\begin{itemize}
\item In the \textbf{early stage}, we observe large variance across samples, even though the mean divergence remains moderate.
This variance reflects the instability of the small model’s predictions. Because this stage governs global structure and motion formation, even small deviations can drastically alter the resulting layout.
Consequently, employing the large model at this stage plays a crucial role in generating high-quality videos.
\item In the \textbf{late stage}, the mean divergence is higher than in the intermediate stage, indicating that the velocity predictions of the small and large models differ most in this stage.
This stage corresponds to fine-grained detail refinement and artifact suppression, both of which critically rely on the representational capacity of the large model.
\end{itemize}

As shown in \fref{fig:velocity_divergence}, we plot the velocity divergence between the small and large models. The empirically determined stage boundaries from \sref{sec:quantitative} are marked with red arrows. Interestingly, for both LTX-Video and WAN~2.1, the stage boundaries empirically determined in \sref{sec:quantitative} coincide with the boundaries inferred from the cosine distance curves between the small and large models. The same pattern holds for the $\ell_2$ distance.
These observations suggest that the characteristic U-shape of velocity divergence could offer a \emph{convenient proxy for identifying stage boundaries}.
In particular, we speculate that the onset of increasing variance in the late stage corresponds to the late stage boundary. Notably, this trend does not appear to be model-specific, metric-specific, or condition-specific. Even unconditioned velocities exhibit a similar behavior (\aref{appdixsec:u_shape}).

Taken together, these findings suggest a simple underlying principle of the LSL schedule introduced earlier:
\textbf{Use the large model when divergence is high and the small model when divergence is low.}

\begin{figure}[t]
  \centering
  \includegraphics[width=\linewidth]{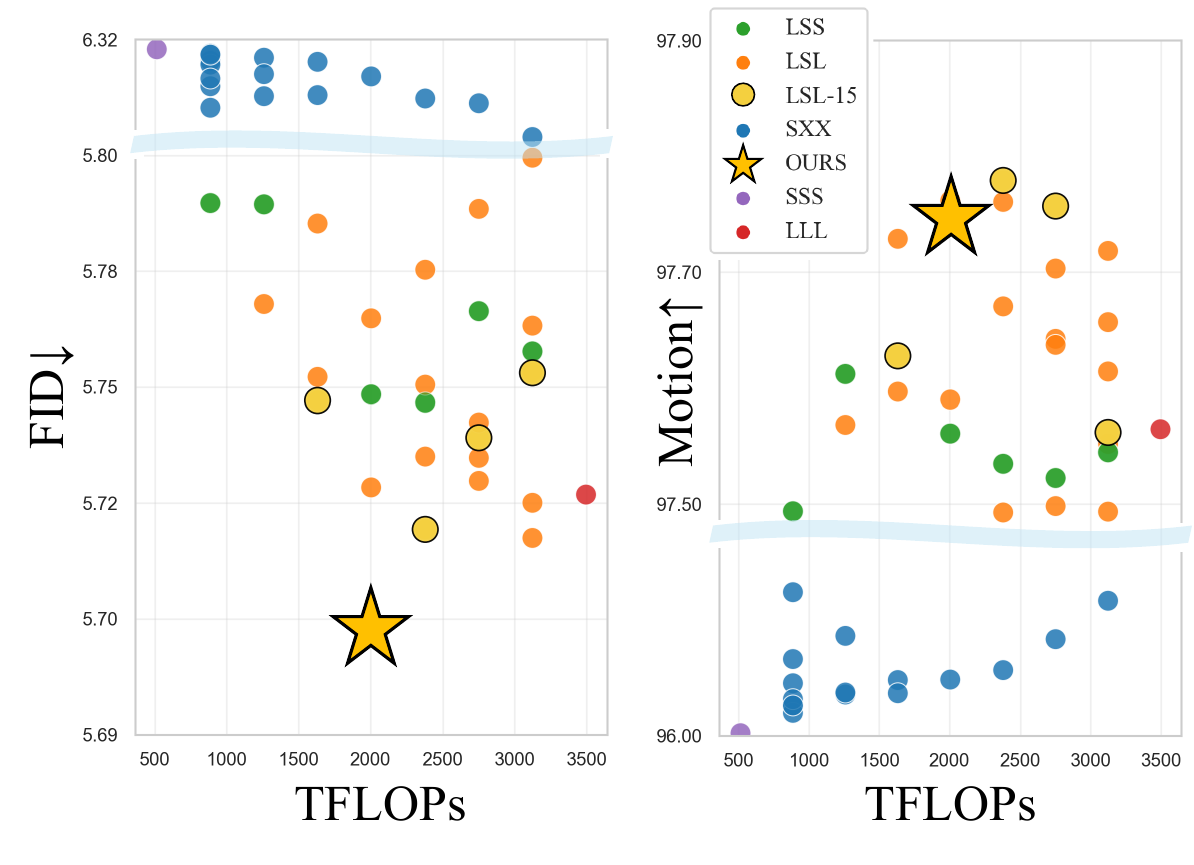}
  \vspace{-1em}
  \caption{
  \textbf{FID vs.\ FLOPs across schedule configurations.}
  Schedules beginning with S (SXX) show significantly degraded quality.
  Schedules using the large model in the early stage (LXX) achieve substantially better performance.
  Our selected early and late boundaries (yellow dots), and especially the full stage-aware schedule (yellow star), lie on the Pareto frontier, achieving LLL-level quality with lower FLOPs.
  }
  \vspace{-1em}
  \label{fig:ablation}
\end{figure}

\subsection{Extensive schedule comparison}
\label{sec:diverse_configurations}
We evaluate nearly the entire space of possible schedules.
To do so, we divide the sampling trajectory into 12.5\% segments and generate combinations of large (L) and small (S) model assignments.
This covers LLL, LSS, LSL, and SSS, along with a broad set of additional S-starting schedules (denoted as SXX in \fref{fig:ablation}) for comparison.
For each schedule, we measure the FID and Motion Smoothness of generated videos.
\fref{fig:ablation} plots FID and Motion Smoothness against computational cost (FLOPs) for the entire schedule configurations.

Consistent with our earlier observations, all schedules that begin with S (blue points in \fref{fig:ablation}) yield significantly worse FID and Motion Smoothness than those beginning with L.
Surprisingly, even the schedule that applies the large model only once at the very beginning (the leftmost green point) performs better than a schedule that applies the large model everywhere except the first segment (the rightmost blue point).
This highlights the crucial role of the early stage in establishing overall video quality.

In \fref{fig:ablation}, the yellow points correspond to LSL schedules that adopt the early stage boundary identified in \sref{sec:early_stage_quantitative}. They generally achieve a more favorable FID–FLOPs trade-off than the orange LSL schedules with alternative early stage boundaries, indirectly supporting the superiority of our boundary choice in \sref{sec:early_stage_quantitative}.

Among these variants, our proposed LSL schedule (marked with a star) attains the lowest FID while preserving an excellent FID–FLOPs trade-off. Despite much lower compute, it delivers consistently strong performance, and qualitatively, its outputs remain nearly indistinguishable from those produced by the large-only baseline (LLL).

\fref{fig:ablation} reveals two additional observations about boundary sensitivity.
(i) Shifted variants, which are the orange points aligned vertically with the yellow star match our FLOPs budget but shift the early or late boundary.
Across both FID and motion smoothness, our configuration (yellow star) consistently outperforms these shifted LSL variants (orange points aligned vertically).

(ii) Expanded/reduced variants, shown as the orange points immediately to the left or right of the yellow star, add or remove a single boundary segment.
These variants (orange points in the vertical band around the yellow star) also yield worse FID and motion smoothness compared to our configuration (yellow star), indicating that even small deviations from the selected boundaries degrade quality.

Together, these results show that, across a wide range of possible configurations, the identified early and late boundaries form a robust sweet spot for capacity allocation.
All quantitative metrics exhibit consistent trends; full tables are included in \aref{appdixsec:all_ablation_fig}.

\section{Compatibility with other acceleration methods}
\label{sef:with_other_acceleration}

In this section, we show that our approach is orthogonal to existing acceleration methods: (a) sampling-step reduction algorithm and (b) distillation-based model training.

\begin{figure}[t]
  \centering
  \includegraphics[width=\linewidth]{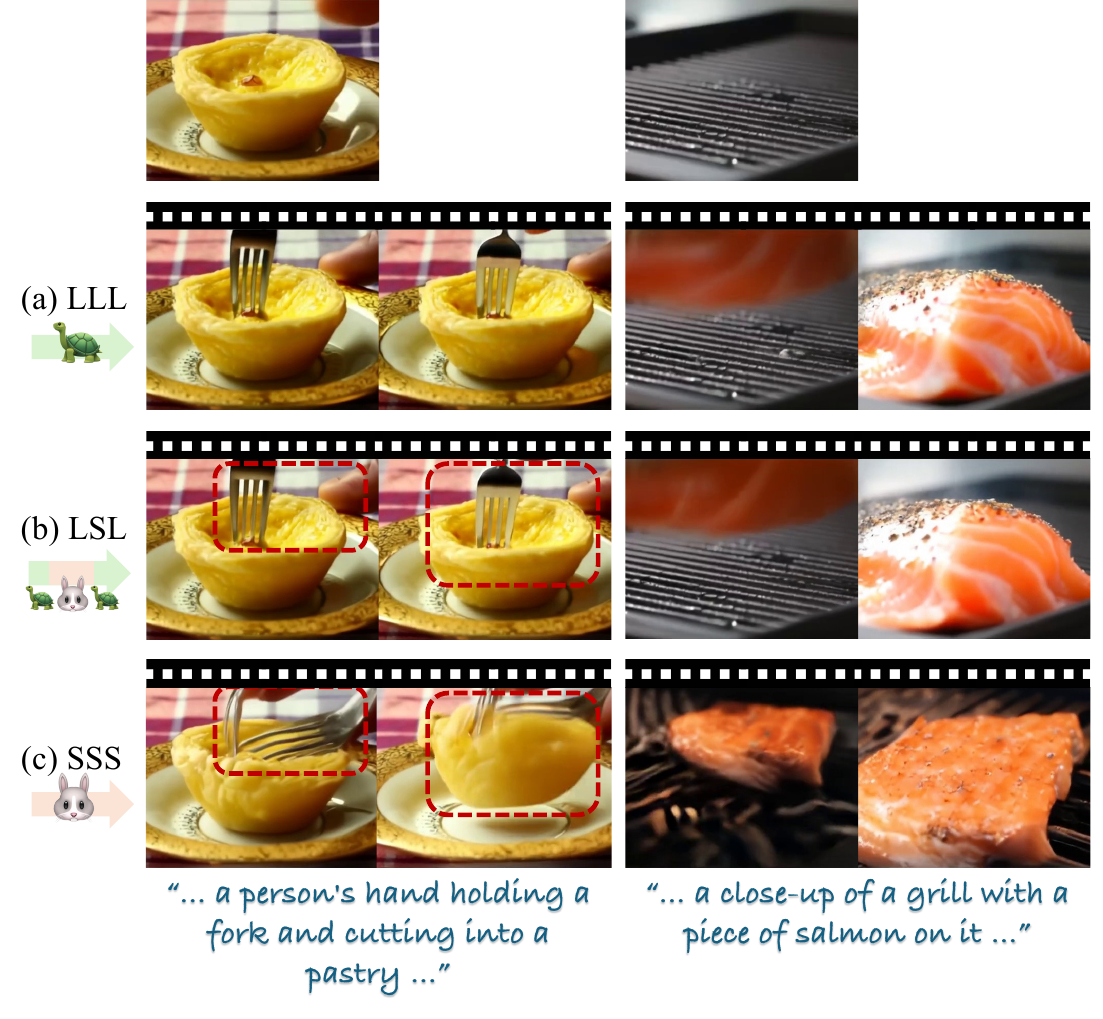}
  \vspace{-2em}
  \caption{
  \textbf{Compatibility with DPM++ solver.} \js{Ours (LSL) is compatible with DPM++ solvers, reproducing similar videos to the videos using only the large model (LLL). Please zoom in to view the figures in detail.}}
  \vspace{-1em}
  \label{fig:sampler}
\end{figure}

\paragraph{(a) Compatibility with sampling-step reduction Algorithm.}
DPM++ accelerates the denoising process by reducing the number of function evaluations (NFE).
We evaluate our LSL schedule together with DPM++ \cite{Lu_2025} with the NFE reduced by half.
As illustrated in \fref{fig:sampler}, even with the reduced NFE, LSL reproduces results comparable to the large-only baseline (LLL). In contrast, the small-only schedule (SSS) produces noticeable artifacts: the fork and the pastry deform (left example), and abrupt scene transitions (right example). This demonstrates that our findings remain valid when combined with DPM++.
Quantitatively, \tref{tab:with_others} shows that DPM++ with LSL reduces total TFLOPs by about $2\times$, with negligible quality loss.

\begin{figure}[t]
  \centering
  \includegraphics[width=\linewidth]{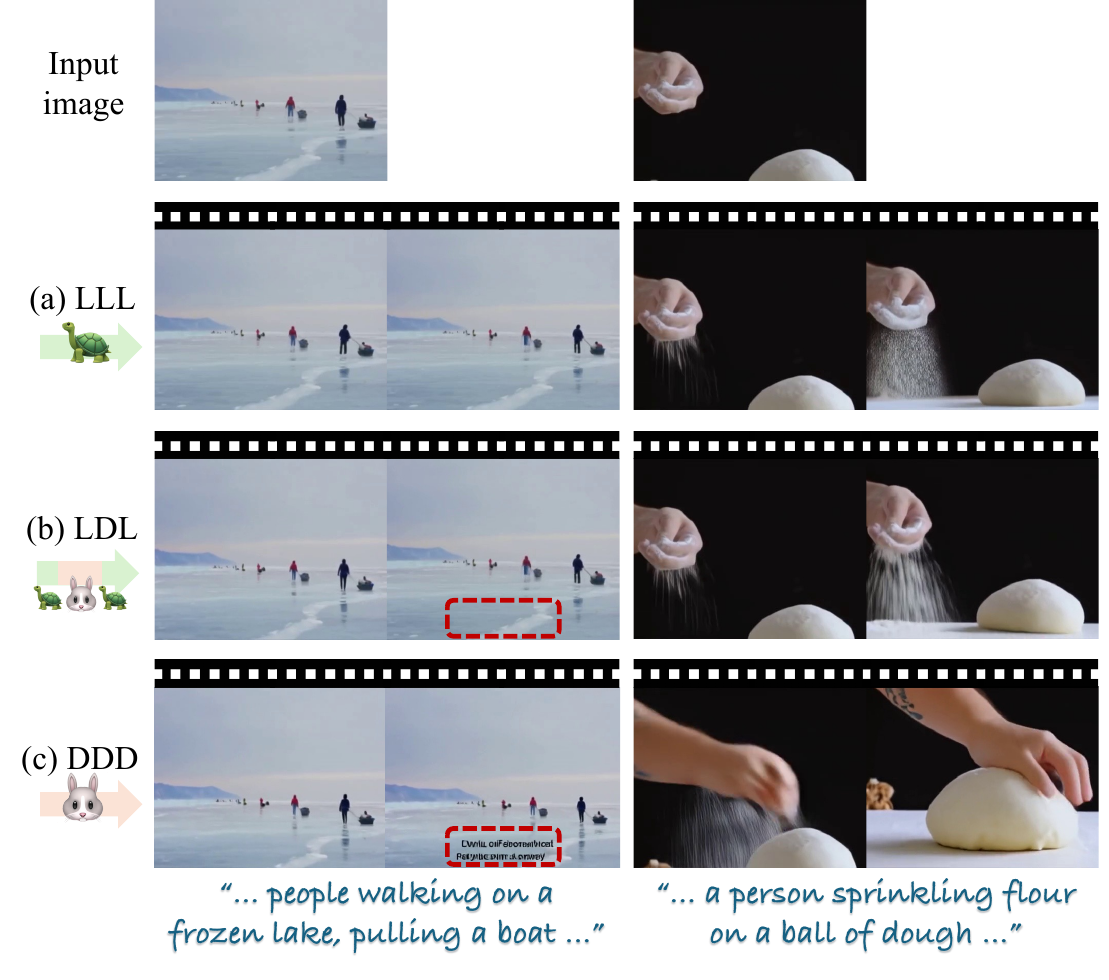}
  \vspace{-2em}
  \caption{
  \textbf{Compatibility with step distillation model.} The step distillation model (D) with small capacity can replace the original small model (S). As done with the original small model, LDL reproduces the results with LLL. In contrast, DDD does not. Please zoom in to view the figures in detail.
  }
  \vspace{-1em}
  \label{fig:distill}
\end{figure}

\paragraph{(b) Compatibility with the distilled model.}
Next, we replace the small model (S) in our LSL schedule with the officially released distilled variant (D) of LTX-Video.
The distilled model was trained to generate videos in just eight sampling steps. Within the LSL framework, we employ it only for the intermediate stage, corresponding to four sampling steps, and denote this configuration as \textbf{LDL}.
As shown in \fref{fig:distill}, LDL reproduces results comparable to the large-only baseline (LLL), avoiding issues such as sudden subtitle appears (left) or hand deformation artifacts (right) that appear in SSS.
\tref{tab:with_others} further confirms that LDL reduces total FLOPs to nearly half of LLL while maintaining comparable perceptual quality.
These results confirm that ours complements existing acceleration techniques and can be \emph{integrated} with them for better efficiency.

\begin{table}[]
\resizebox{\columnwidth}{!}{%
\begin{tabular}{lllcccc}
\toprule
  \textbf{Setting}                        
&\textbf{Solver}& \textbf{Schedule}  &\textbf{NFE}& \textbf{FID $\downarrow$} & \textbf{FVD $\downarrow$} & \textbf{TFLOPs $\downarrow$}\\ 
\midrule
  \multirow{3}{5.95em}{Sampling-Step Reduction}&\multirow{3}{*}{DPM++}& LLL               
 &20& \underline{5.78}                      & \textbf{769.83}                    &                               1748
\\
  
&& LSL               
 &20& \textbf{5.77}                      & \underline{773.32}                    &                               928
\\
                                          
&& SSS                &20& 6.11                      & 1,023.65                  &                               257\\
\midrule
 \multirow{3}{*}{Distillation}
& \multirow{3}{*}{ODE}& LLL                &40& \textbf{5.73}                      & \textbf{834.26}                    &3496 
\\
 & & LDL &24& \underline{5.93}& \underline{1,104.18}&1774
\\
 & & DDD &8& 6.50                      & 1,843.53                  &51\\
\bottomrule
\end{tabular}%
}
\caption{
\textbf{Compatibility with existing acceleration methods.}
Our stage-aware schedule maintains large-model quality even when combined with reduced-step solvers or distilled backbones, demonstrating orthogonality to existing accelerators.}
\vspace{-1em}
\label{tab:with_others}
\end{table}

\section{Discussion and conclusion}
\label{sec:discussion}
We investigate how model capacity contributes differently across the video diffusion denoising process.
Through empirical analyses, we show that the importance of model capacity varies across denoising stages:
The early stage governs global structure and motion, and the late stage refines high-frequency details and suppresses artifacts, both of which are capacity-sensitive.

Building on this insight, we introduce FlowBlending, a stage-aware multi-model sampling strategy that allocates the large model to the early and late stages while delegating the intermediate stage to the small model.
FlowBlending preserves the visual fidelity, temporal coherence, and semantic alignment of the large model, yet notably reduces sampling cost.
Across LTX-Video and WAN 2.1, \ours{} achieves up to \textbf{1.65× faster} inference with \textbf{57.35\% FLOPs} while not compromising the video quality.

To determine where model capacity matters most, we proposed a practical method for identifying the early and late stage boundaries using DINO similarity and FID-based trade-off analysis. Interestingly, these empirically chosen boundaries coincide with regions of increasing variance in the velocity divergence between the large and small models. 

With the analysis, we showed that FlowBlending is orthogonal to existing acceleration techniques. In particular, we empirically observe that DPM++ solver produces outputs that more closely match those of the large model. It is another promising research avenue to determine whether improved solvers shift the effective stage boundaries.

While FlowBlending is broadly applicable, a key limitation remains: stage boundaries often need to be re-estimated when the diffusion model changes. Automatic boundary detection or model-agnostic stage criteria would further improve the usability and generality of stage-aware sampling.




{
    \small
    \bibliographystyle{ieeenat_fullname}
    \bibliography{main}
}

\clearpage
\appendix
\setcounter{page}{1}
\maketitlesupplementary

\renewcommand{\thetable}{A\arabic{table}}
\renewcommand{\thefigure}{A\arabic{figure}}
\setcounter{figure}{0}
\setcounter{table}{0}
\renewcommand{\theHtable}{A\arabic{table}}
\renewcommand{\theHfigure}{A\arabic{figure}}

\section{Experimental Details}
\label{appdixsec:exp_setting}

We evaluate the proposed sampling schedule on two representative open-source video diffusion models:
LTX-Video (2B / 13B) \cite{hacohen2024ltx} and WAN 2.1 (1.3B / 14B) \cite{wan2025wan}.
LTX-Video generates 65-frame videos at a resolution of 608$\times$342 and 30 fps on the VBench dataset \cite{huang2024vbench}, and 832$\times$480 videos on the PVD \cite{bolya2025perception}.
WAN 2.1 generates 61-frame videos at 832$\times$480 and 16 fps across both datasets.

For quantitative evaluation, we report FID \cite{fid} and FVD \cite{fvd} on 284 generated samples using subset of PVD prompts, and VBench metrics on 355 generated samples using VBench dataset prompts or 284 generated samples same as FID and FVD.
We use the default evaluation configurations provided by each official repository: LTX-Video experiments adopt the ODE sampler, while WAN 2.1 experiments follow the UniPC sampler.

To assess computational efficiency, we additionally measure runtime and compute FLOPs of DiT blocks per generated video.
LTX-Video runtime is measured on an NVIDIA A6000 GPU, and WAN 2.1 runtime is measured on an NVIDIA A100 GPU.

Most of our quantitative results, including FID, FVD, and qualitative comparisons, are obtained using videos generated on the PE Video Dataset, which we adopt as our primary evaluation set. Although we also report VBench metrics, the results are rely on PVD due to its richer prompts and better alignment with the types of video quality differences we aim to study.

VBench includes a large collection of short and simple prompts designed to probe a wide range of video properties.
However, many of these prompts produce videos that do not align with the types of content we aim to evaluate.
For instance, prompts such as “a leafless tree”, “young guy with VR headset”, or “milk and cinnamon rolls” typically result in highly static scenes with minimal motion or structural complexity.
Such samples make it difficult for VBench metrics to capture the aspects we care about most object consistency, motion coherence, and subtle high-frequency artifacts.

In contrast, the PE Video Dataset provides substantially richer text–video pairs with more dynamic and semantically meaningful prompts (e.g., “a dog running across a snowy field,”).
Evaluations with these prompts better reflect the perceptual differences we observe empirically across sampling schedules.
While both VBench and PE yield broadly similar trends, PE offers a closer match to our practical experience regarding model quality, making it our primary benchmark dataset.

\section{More ablation}
\label{appdixsec:more_abl}
\subsection{Early stage ablation}
\label{appdixsec:more_abl_early}
\paragraph{Experimental details}

To determine the early stage boundary, we conduct additional ablation experiments using LTX-Video (2B / 13B), an image-to-video model, and WAN 2.1 (1.3B / 14B), a text-to-video model.
We follow the default configurations provided by the official repositories, including the sampler, noise schedule, and total number of denoising steps.
In detail, we use 40 timesteps for LTX-Video and 50 timesteps for WAN 2.1.

For evaluation, we generate 355 videos using the text and image prompts from the VBench Dataset \cite{huang2024vbench}.
To quantify how well the small model can replace the large model at different early stages, we measure frame-wise DINO similarity, CLIP similarity, LPIPS, PSNR, and SSIM between each hybrid schedule that switches from the large to the small model ($L \rightarrow S$) at a given timestep and the large-only baseline (LLL).
Each metric is computed between corresponding frames of the hybrid schedule and the LLL output, and then averaged across all frames and videos:

\begin{equation}
\text{Metric}(t_{\text{switch}}) =
\frac{1}{N}\sum_{i=1}^{N}
d\big(x^{(L\rightarrow S)}_{i},\; x^{(L)}_{i}\big),
\end{equation}

where $x^{(L)}_{i}$ denotes the $i$-th frame of the large-only baseline.

We progressively shift the early stage boundary across timesteps:
LTX-Video experiments vary the boundary in increments of 5 timesteps, and
WAN 2.1 experiments vary it in increments of 10 timesteps.
This allows us to precisely characterize how sensitive the early denoising stage is to model capacity across both architectures.

\paragraph{Results}

\begin{figure}[t]
  \centering
  \includegraphics[width=\linewidth]{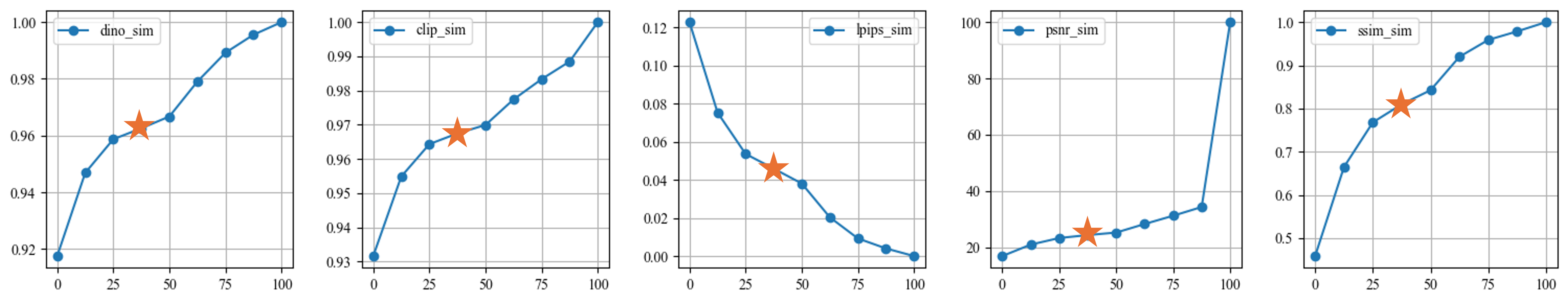}
  \caption{
  \textbf{LTX-Video: Experiments for choosing early stage boundary }. 
  The star sign represents our choice of the early stage boundary. The X-axis is the small model introduction point (\%), where the small model follows the large model. We present the results for CLIP, LPIPs, PSNR, and SSIM in order.
  }
  \label{fig:more_abl_early_ltx}
\end{figure}

\begin{figure}[t]
  \centering
  \includegraphics[width=\linewidth]{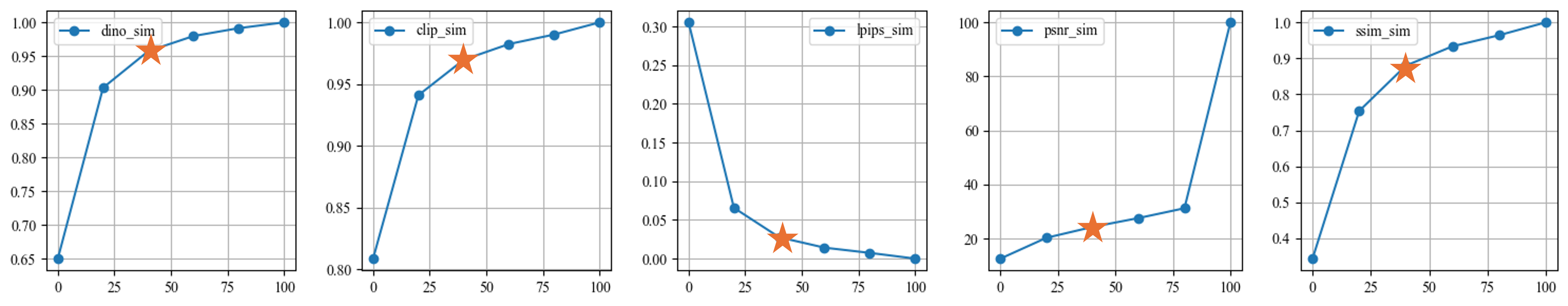} 
  \caption{
  \textbf{WAN 2.1: Experiments for choosing early stage boundary.} 
  The star sign represents our choice of the early stage boundary. The X-axis is the small model introduction point (\%), where the small model follows the large model. We present the results for CLIP, LPIPs, PSNR, and SSIM in order.
  }
  \label{fig:more_abl_early_wan}
\end{figure}

\fref{fig:more_abl_early_ltx} and \fref{fig:more_abl_early_wan} plot the early stage boundary on the x-axis, expressed as the Small-model introduction point (\%) measured from the start of the sampling steps. The y-axis reports the similarity between the outputs of the large-only model and those of the LSS schedule, evaluated using DINO, CLIP, LPIPS, PSNR, and SSIM. These curves indicate how closely the results resemble the large-only baseline as the early stage boundary varies. As shown in the figures, not only the DINO similarity discussed in the main text but all metrics exhibit a consistent trend: a sharp decline in similarity emerges beyond a specific point in the curve. We select the boundary just before this drop, which preserves global structure and motion.

Among all metrics, we find that DINO and CLIP similarity, both of which capture richer semantic information, align most closely with our observations. This is likely because small pixel-level differences measured by metrics such as LPIPS, PSNR, or SSIM do not always correspond to perceptually meaningful changes.

\paragraph{More qualitative results}
Please see the attached HTML file (“index.html”) for the ablation study on selecting the early-stage boundary.
We provide videos generated with different small-model introduction points (\%) in the denoising process, as illustrated in \fref{fig:dino_graph}.
A setting of 40\% (LSL) means that the small model is introduced starting from 40\% of the denoising process, and 100\% (LLL) corresponds to not introducing the small model at all.
When the introduction point is delayed up to 40\%, the resulting appearance and motion remain nearly identical to the large-only model (LLL). However, pushing the boundary further to 60\% (LSS) leads to a different motion.
Therefore, our final choice, introducing the small model at 40\%, provides the optimal balance between quality and efficiency.

\subsection{Late stage ablation}
\label{appdixsec:more_abl_late}
\paragraph{Experimental details}
This section provides additional information on the setup used to determine the late stage boundary. We experiment with two video diffusion families: LTX-Video (2B/13B), an image-to-video model, and WAN 2.1 (1.3B/13B), a text-to-video model, and generate 284 videos using the PVD dataset for evaluation. We follow all default settings from each official repository, including the sampler configuration and the total number of timesteps, where LTX-Video uses 40 steps and WAN 2.1 uses 50 steps. In all experiments, we fix the early stage boundary identified in Section 4.2 and vary only the late stage boundary. For each schedule, we measure performance using VBench metrics—Subject Consistency, Background Consistency, Temporal Flickering, Motion Smoothness, Dynamic Degree, Aesthetic Quality, and Image Quality—along with FID and FVD. Ablations are performed by gradually shifting the late stage boundary forward from the fixed early boundary, using 5-timestep intervals for LTX-Video and 10-timestep intervals for WAN 2.1. All LTX-Video results are obtained on an NVIDIA A6000 Ada GPU, while WAN 2.1 evaluations are conducted on an NVIDIA A100 GPU due to model size differences.

\begin{figure}[t]
  \centering
  \includegraphics[width=\linewidth]{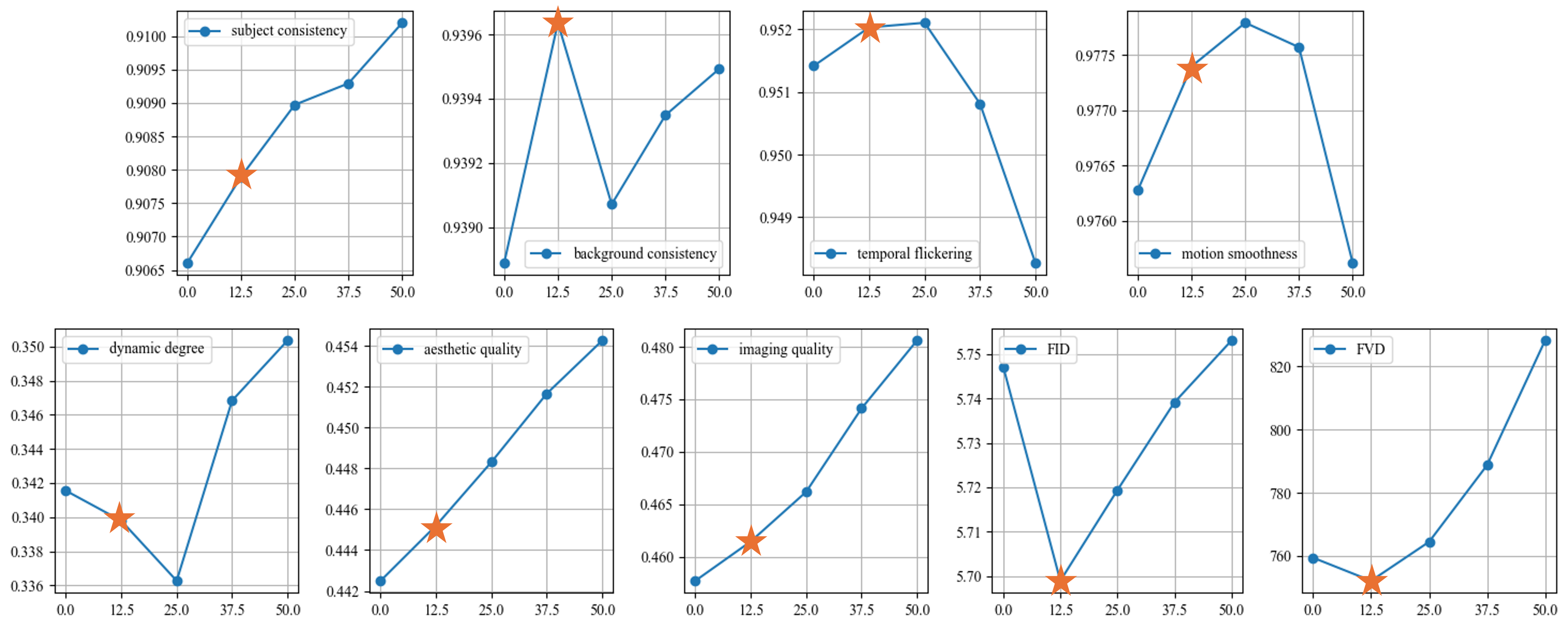}
  \caption{
  \textbf{LTX-Video: Experiments for choosing late stage boundary.} The star sign represents our choice of the late stage boundary. The star sign represents the large model reintroduction points (\% from end). We measure Subject Consistency, Background Consistency, Temporal Flickering, Motion Smoothness, Dynamic Degree, Aesthetic Quality, Image Quality, FID, and FVD.
 }
  \label{fig:more_abl_late_ltx}
\end{figure}

\paragraph{Results}
\label{appendix:late_stage_ablation_quanti}
In \fref{fig:more_abl_late_ltx}, we provide the results of LTX-Video with the Vbench metrics, FID, and FVD. The star sign represents our late boundary choice, and the optimal choice of the late boundary enables us to maintain the video quality.
As shown in the \fref{fig:more_abl_late_ltx}, FID and FVD graphs have the best scores with our late boundary choice. With our late boundary choice, the video quality-related metrics and background consistency also show the best and comparable results, respectively. Likewise, the temporal flickering and motion smoothness achieves nearly on par with the best results. 
Although our late boundary choice always work on the other metrics, their measurements are far from our purpose of using the big model at the late stage.
For example, measuring the dynamic degree and aesthetic score is irrelevant to maintaining fine details of the video. In the imaging quality metric, using more big model results in a better score. However, the observed visual difference is negligible.

\begin{figure}[t]
  \centering
  \includegraphics[width=\linewidth]{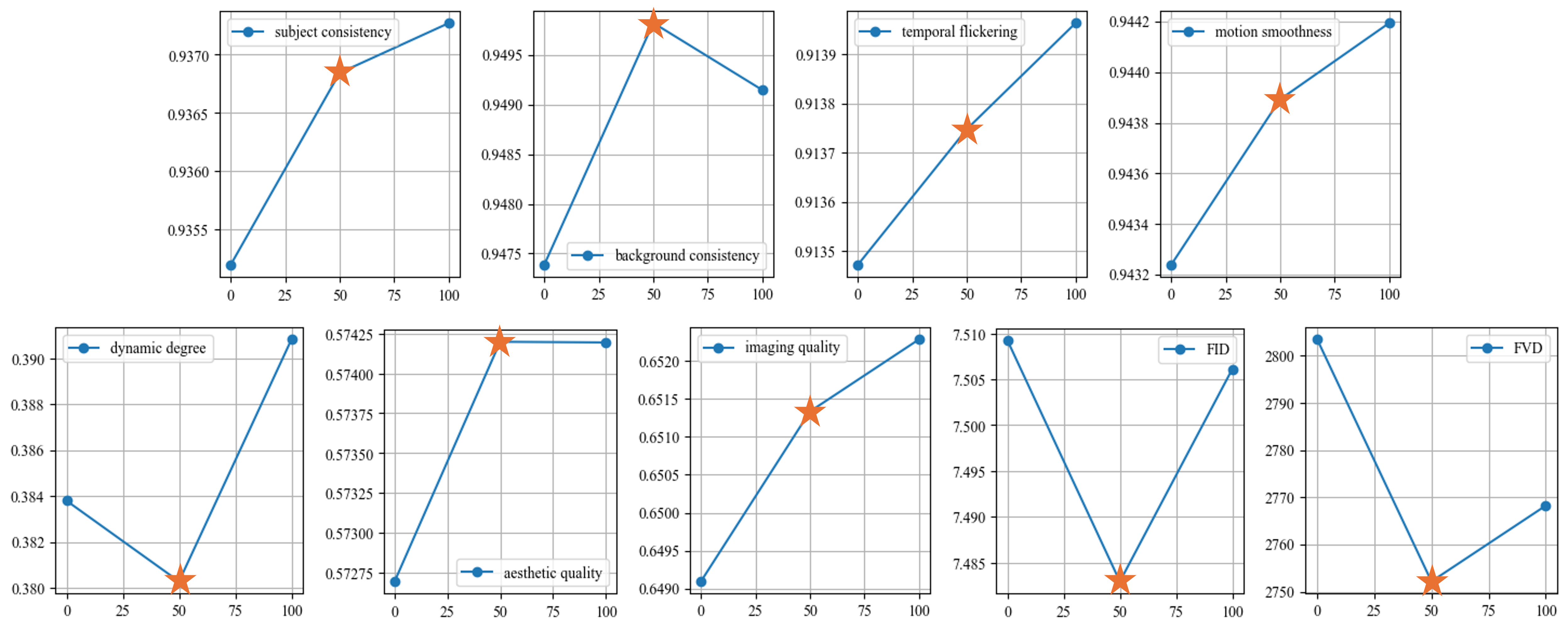}
  \caption{
  \textbf{WAN 2.1: Experiments for choosing late stage boundary.} The star sign represents our choice of the late stage boundary. The star sign represents the large model reintroduction points (\% from end). We measure Subject Consistency, Background Consistency, Temporal Flickering, Motion Smoothness, Dynamic Degree, Aesthetic Quality, Image Quality, FID, and FVD.
}
  \label{fig:more_abl_late_wan}
\end{figure}

In \fref{fig:more_abl_late_wan}, we provide the Vbench metrics, FID, and FVD. Similar to the LTX-Video setting, we can identify an optimal late boundary by locating the points where FID and FVD are minimized. 
Moreover, the overall trends are consistent with those observed for LTX-Video across different metrics, indicating that our late boundary strategy behaves robustly regardless of the specific evaluation metric.

In Algorithm \ref{algo:knee}, we provide the pseudo-algorithm to find the early stage boundary with the slope. We set the threshold to find the late stage boundary with FID.

\begin{algorithm}[t]
\caption{How to find early stage boundary via relative slope threshold}
\label{algo:knee}
\begin{algorithmic}[1]
\STATE \textbf{Input:} The small model introduction points (\%) $\{x_i\}_{i=0}^{N-1}$, DINO similarities $\{y_i\}_{i=0}^{N-1}$, threshold $\alpha \in (0,1)$
\STATE \textbf{Output:} Early stage boundary index $k$, Early stage boundary location $(x_k, y_k)$

\STATE \textbf{Step 1: Compute local slopes}
\FOR{$i = 0$ to $N-2$}
    \STATE $\Delta x \gets x_{i+1} - x_i$
    \IF{$\Delta x = 0$}
        \STATE $s_i \gets 0$
    \ELSE
        \STATE $s_i \gets (y_{i+1} - y_i) / \Delta x$
    \ENDIF
\ENDFOR

\STATE \textbf{Step 2: Reference slope}
\STATE $s_{\mathrm{ref}} \gets s_0$

\STATE \textbf{Step 3: Find index where slope falls below threshold}
\STATE $k \gets N - 1$
\FOR{$i = 0$ to $N-2$}
    \IF{$s_i \le \alpha \, s_{\mathrm{ref}}$}
        \STATE$k \gets i + 1$
        \STATE \textbf{break}
    \ENDIF
\ENDFOR

\STATE \RETURN $(k, x_k, y_k)$

\end{algorithmic}
\end{algorithm}

\paragraph{Qualitative results}
\label{appendix:late_stage_ablation_qual}

Please see the attached HTML file (“index.html”) for the ablation study on selecting the late-stage boundary.
Therefore, reintroducing the large model at 20\%, our final choice, offers the optimal balance between video quality and efficiency.
We provide videos generated with different large-model reintroduction points (\% from the end) in the denoising process, as illustrated in \fref{fig:fid_curve}.
A setting of 40\% (LSL) means that the large model is reintroduced starting from the last 40\% of the denoising trajectory; 20\% (LSL-ours) reintroduces the large model from the last 20\%; and 0\% (LSS) corresponds to not reintroducing the large model.
When the reintroduction point is set to 20\%, the model preserves fine-grained textures without introducing noticeable artifacts.
However, pushing the boundary further to 0\% (LSS) introduces visible artifacts-most prominently, a severe artifact on the dog’s eye in the last frame.
Therefore, reintroducing the large model at 20\%, our final choice, offers the optimal balance between video quality and efficiency.

\section{U-shaped divergence curve}
\label{appdixsec:u_shape}

\subsection{Experimental details}

We measured the divergence between the velocity fields of the small and large models across timesteps.

\begin{figure}[t]
  \centering
  \includegraphics[width=0.5\linewidth]{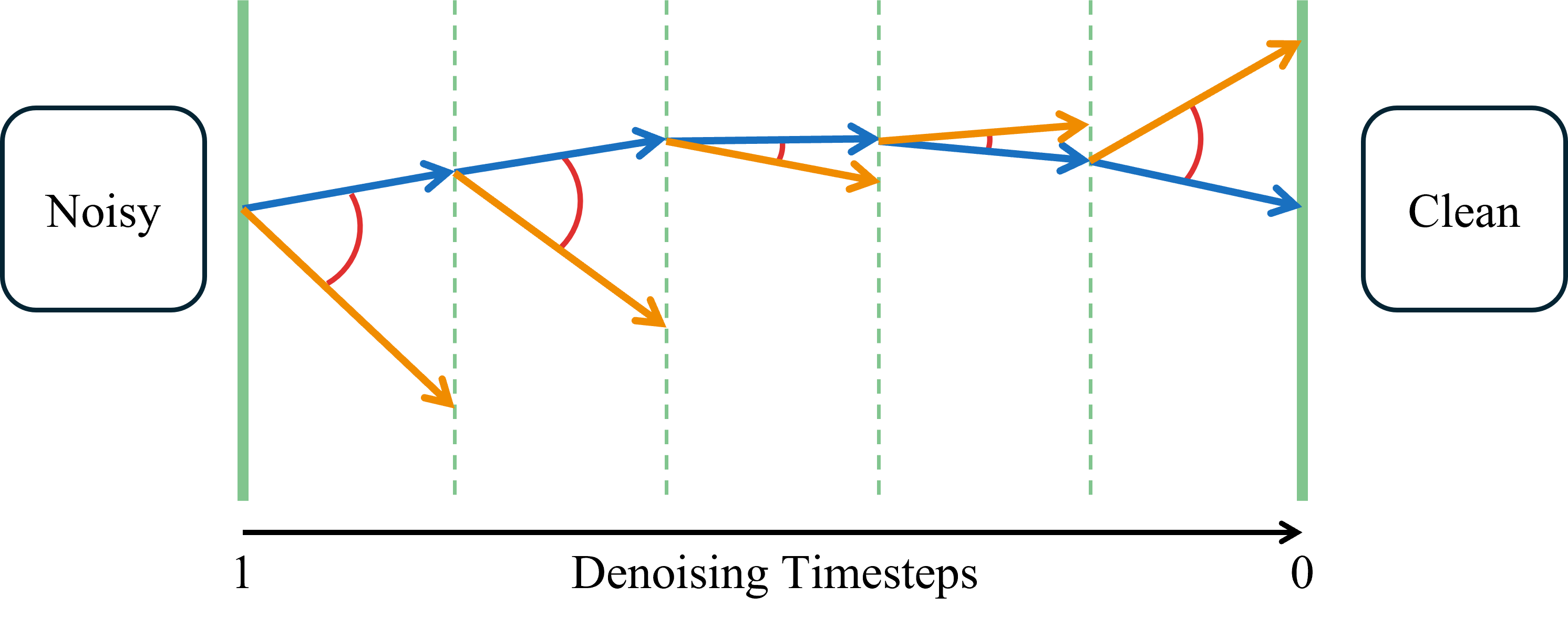}
  \caption{
  \textbf{Illustration of velocity divergence across diffusion steps.}
  We follow the sampling trajectory of the main-stream model (blue) and feed its intermediate outputs into both the large and small models at each timestep. 
The divergence between their velocity predictions is then computed using cosine distance and L2 distance. 
This procedure is applied identically to LTX-Video and WAN2.1, using the official inference settings of each model.
}
  \label{fig:u_shape_explain}
\end{figure}

The details are as follows.
LTX-Video \cite{hacohen2024ltx} is an image-to-video (I2V) model, and Wan2.1 (1.3B/13B) \cite{wan2025wan} is a text-to-video (T2V) model.
We measured 355 videos using the text and image prompt from VBench dataset \cite{huang2024vbench}.
We use the default sampler and inference timesteps, following the official repository of each video model. We measured the runtime of LTX-video and WAN2.1 on a NVIDIA A6000 GPU and on a NVIDIA A100 GPU, respectively. In the \fref{fig:u_shape_explain}, we provide the illustration of how we measure the divergence between the velocity fields of the small and large models across timesteps. As shown in the figure, we set the sampling trajectory of the large models as a \textit{main stream}, and we fed the previous large model outputs to the large model and the small model during inference.  We measure the divergence between the velocity predictions of the small and large models at each time step. We use cosine distance and L2 distance as measurements.

\begin{figure}[t]
  \centering
  \includegraphics[width=\linewidth]{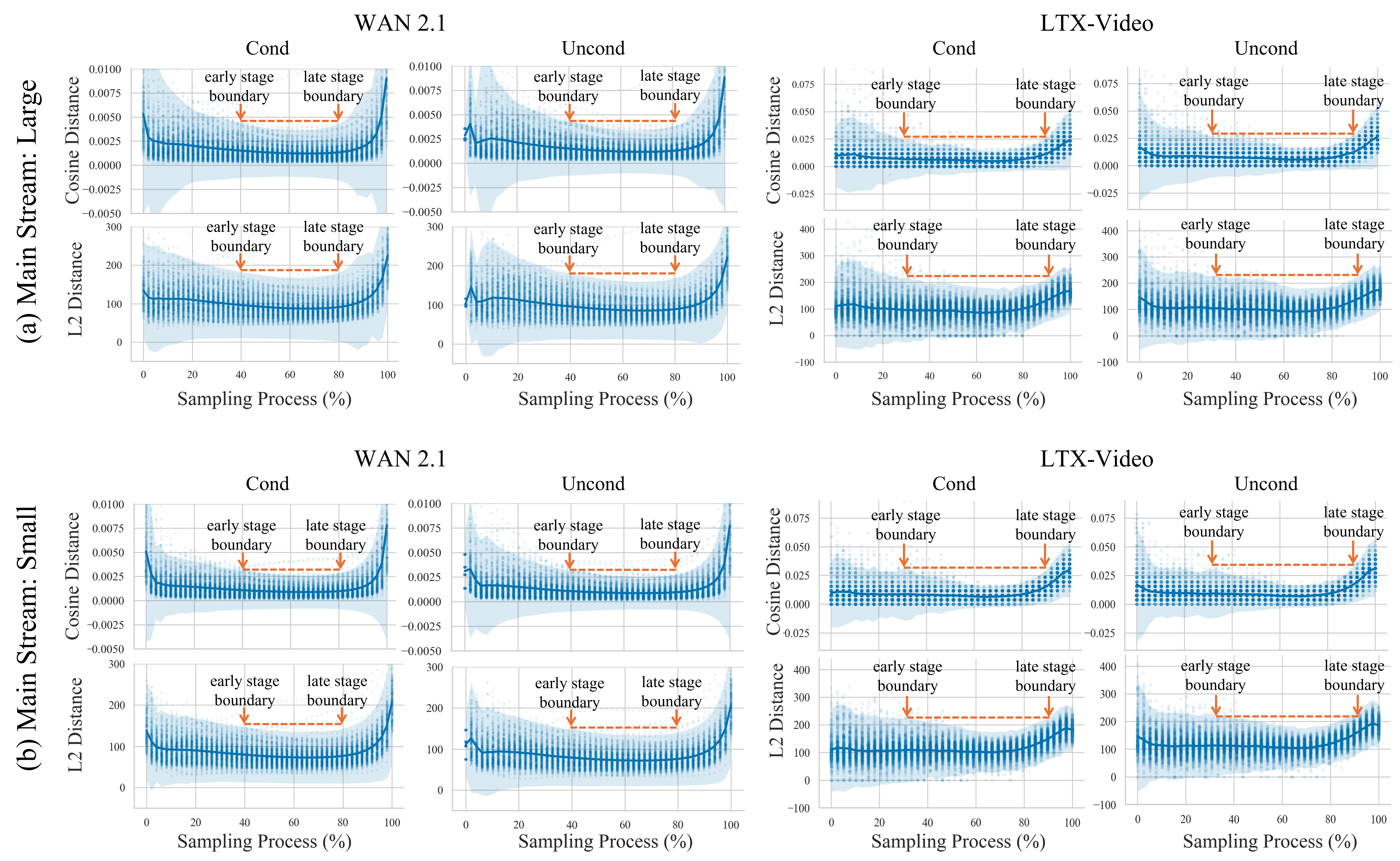}
  \caption{
  \textbf{Velocity divergence across diffusion steps.} 
  We measure the divergence between the velocity predictions of the small and large models at every denoising step for LTX-Video and WAN2.1. 
  (a) uses the large model as the main stream, while (b) uses the small model as the main stream. 
  For each timestep, we compare the velocity predicted with the conditional input (cond) and the null input (uncond), corresponding to the CFG-conditioned and unconditioned branches, respectively. 
  Divergence is lowest in the intermediate stage, indicating that the small model denoises reliably there, whereas the early and late stages show higher divergence due to structure formation and detail refinement.
  }
  \label{fig:u_shape}
\end{figure}

In the \fref{fig:u_shape} (a), we provide the experiment results. Divergence is lowest in the intermediate stage, indicating that the small model can reliably denoise there. Early stage exhibits high variance (structure formation), while late stage
shows high mean divergence (detail refinement). This supports allocating the large model to the early and late stages. Interestingly, we found a similar result with the small model as a mainstream.

\section{Comprehensive schedule sweep}
\label{appdixsec:all_ablation_fig}
\begin{figure}[t]
  \centering
  \includegraphics[width=0.9\linewidth]{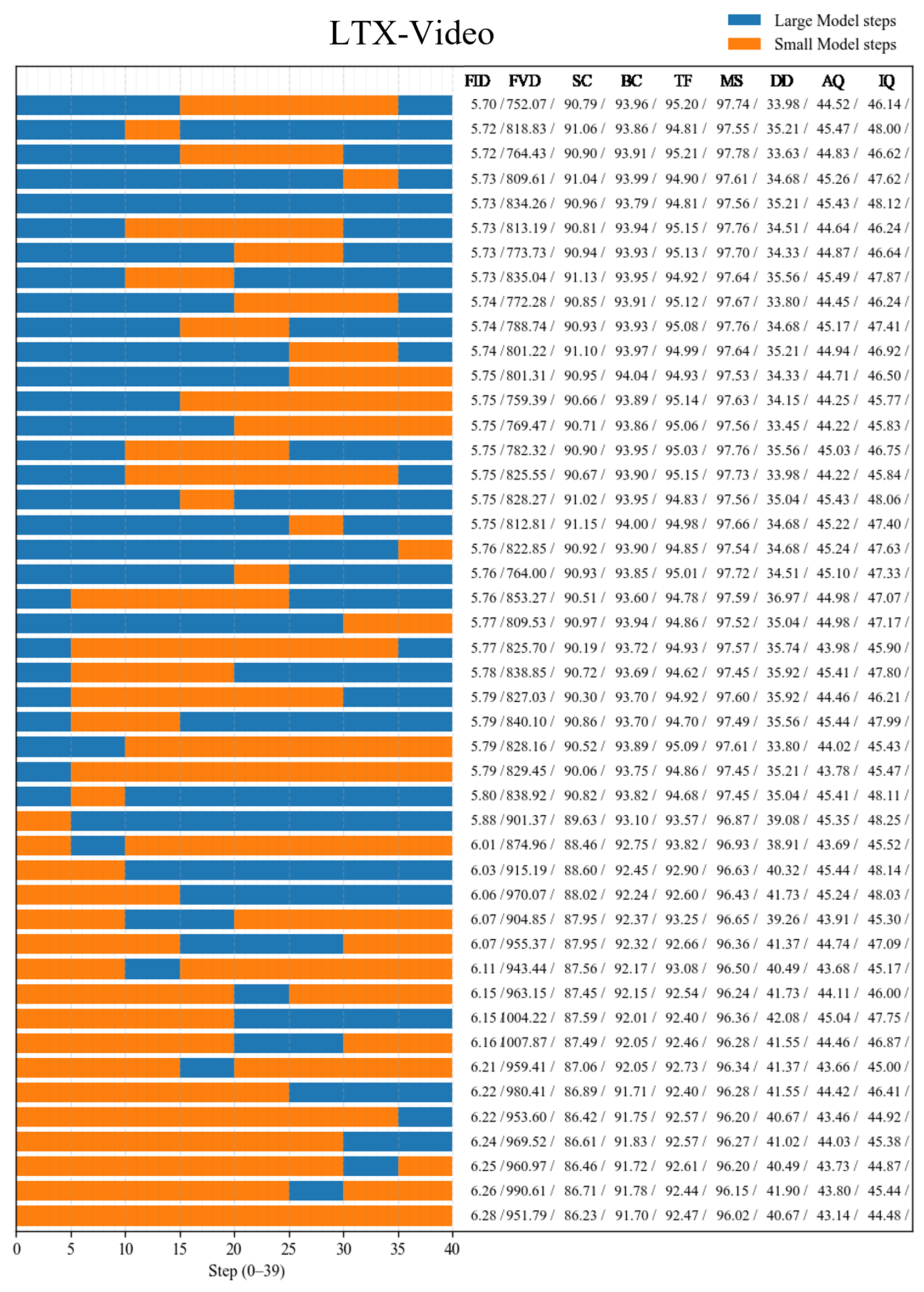}
  \caption{
  \textbf{LTX-Video: Comprehensive schedule sweep on the PV dataset.} 
  We evaluate all combinations of small–large scheduling for LTX-Video on the PV dataset and sort the results in ascending order of FID. Our LSL configuration consistently ranks near the top, demonstrating strong performance across both FID and FVD. Other metrics also exhibit a similar overall trend.
  }
  \label{fig:all_ablation_fig_pe_ltx}
\end{figure}

\begin{figure}[t]
  \centering
  \includegraphics[width=0.9\linewidth]{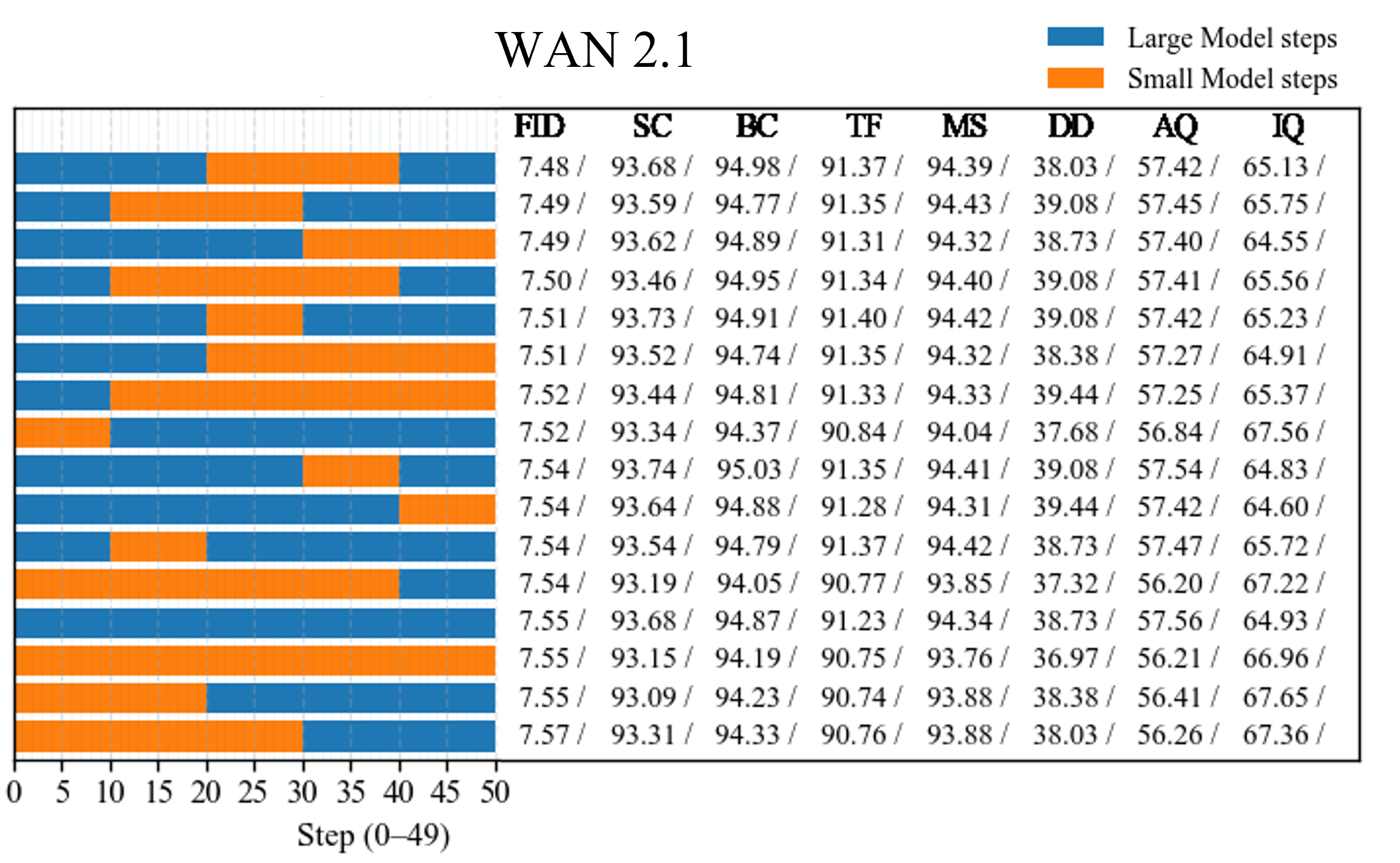}
  \caption{
\textbf{WAN 2.1: Comprehensive schedule sweep on the PV dataset.}
We apply the same exhaustive combination sweep to WAN 2.1 and sort all configurations by FID. Similar to LTX-Video, the LSL strategy consistently outperforms other schedules across both FID. Other metrics also exhibit a similar overall trend.
}
  \label{fig:all_ablation_fig_pe_wan}
\end{figure}

We further provide the evaluations using FID, FVD, and VBench across various combinations of the small and large models.
In \fref{fig:all_ablation_fig_pe_ltx}, we evaluate LTX-Video \cite{hacohen2024ltx} on the PV dataset \cite{bolya2025perception} and sort all combinations in ascending order of their FID scores. We observe that our LSL strategy consistently outperforms the other combinations.
In \fref{fig:all_ablation_fig_pe_wan}, we present the results on the VBench dataset using the same combination order obtained from the PV dataset.

\begin{figure}[t]
  \centering
  \includegraphics[width=0.9\linewidth]{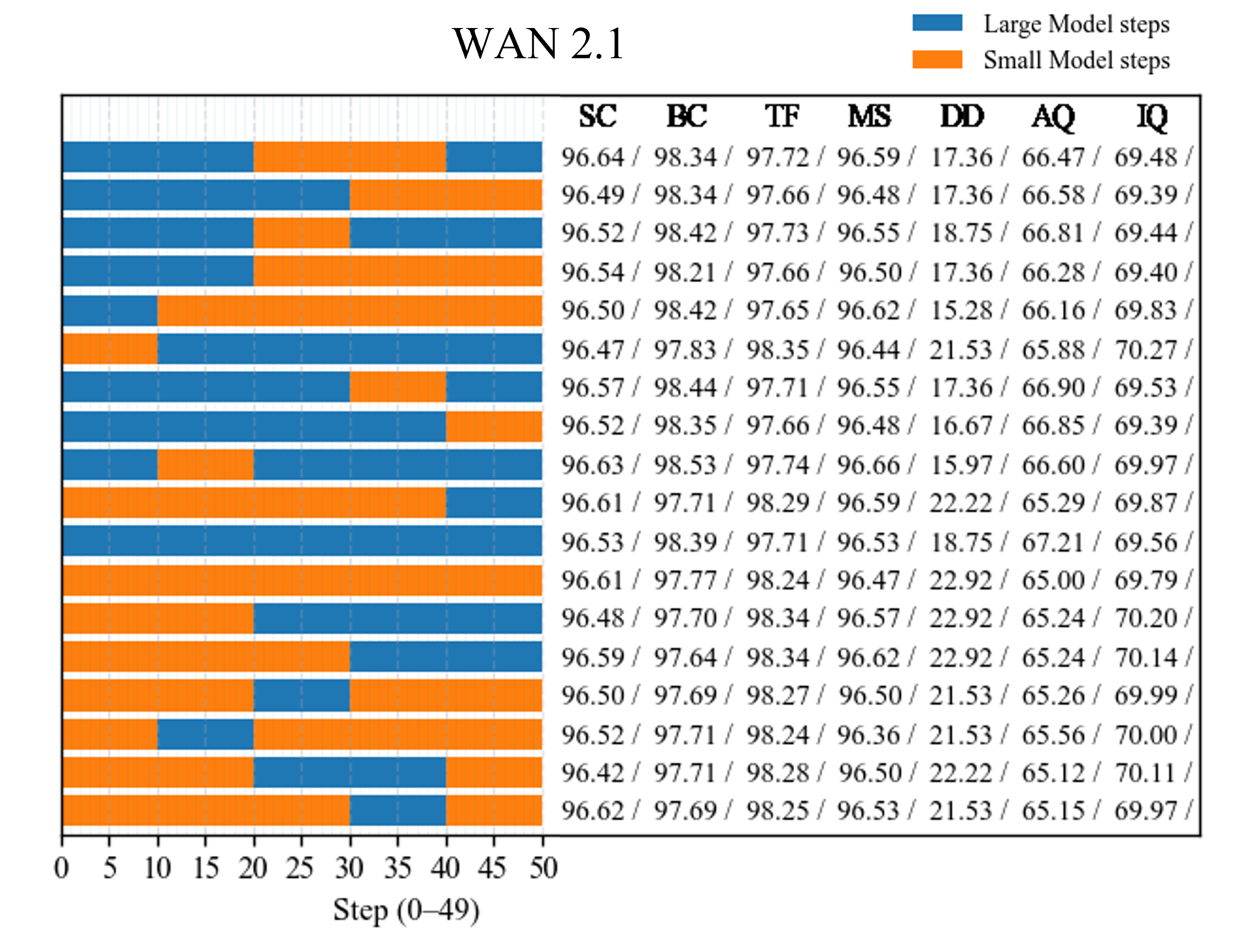}
    \caption{
    \textbf{WAN 2.1: VBench evaluation under the same schedule ordering.}
    Using the schedule ordering obtained from the PV dataset, we report WAN 2.1 performance on VBench. The LSL schedule again provides strong performance across multiple VBench dimensions, confirming its general effectiveness.
    }
  \label{fig:all_ablation_fig_vbench_wan}
\end{figure}

Likewise, we provide the evaluation results with WAN2.1 \cite{wan2025wan}.
In \fref{fig:all_ablation_fig_pe_wan}, we evaluate Wan2.1 on the PV dataset and sort all combinations in ascending order of their FID scores. We also observe that our LSL strategy consistently outperforms the other combinations. In \fref{fig:all_ablation_fig_vbench_wan}, we present the results on the VBench dataset using the same combination order obtained from the PV dataset.

\begin{figure}[t]
  \centering
  \includegraphics[width=0.9\linewidth]{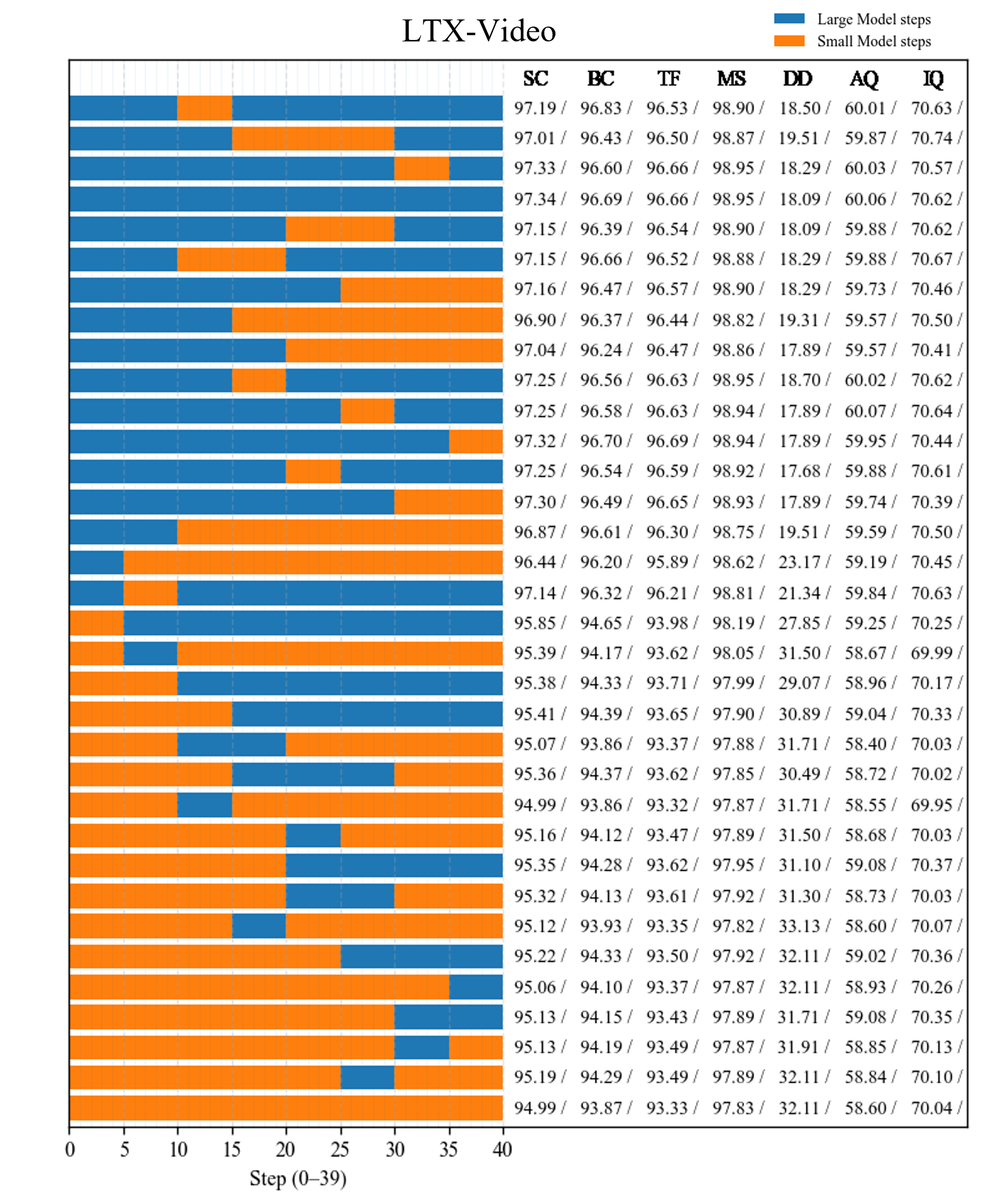}
  \caption{
\textbf{LTX-Video: VBench evaluation under the same schedule ordering.}
Using the ordering derived from the PV dataset sweep, we evaluate LTX-Video on VBench metrics. The relative ranking remains consistent, and the LSL configuration again shows robust performance across diverse quality dimensions. Other metrics also exhibit a similar overall trend.
}
  \label{fig:all_ablation_fig_vbench_ltx}
\end{figure}

\section{Qualitative results}
\label{appdixsec:more_quali}
We include the videos corresponding to all figures presented in the main paper in the supplementary material. 
Please refer to our project page (\url{https://jibin86.github.io/flowblending_project_page/})

\section{More Related Work}

Beyond the methods introduced in the main paper, several additional approaches have been proposed to accelerate diffusion sampling. Adversarial distillation has been explored as an effective way to reduce the number of inference steps \cite{xu2024ufogen, sauer2024adversarial, sauer2024fast}. Based on a deeper understanding of diffusion dynamics, training-free acceleration techniques have also been developed, leveraging step-wise feature reuse \cite{liu2025timestep, lv2024fastercache, zhang2025training}.

\citet{pan2024t} is the most closely related to our work, as they also adopt different models for different denoising stages. However, unlike \citet{pan2024t}, whose aim is primarily to improve image quality, our goal is to preserve the fidelity of a large video diffusion model while reducing computational cost. In video generation, early-stage misalignment in motion cannot be corrected by a small model in later stages. To tackle this challenge, we analyze the stage-specific behavior of video diffusion models and introduce a stage-aware strategy that determines stage boundaries using capacity-sensitive metrics.

\end{document}